\documentclass[11pt,a4paper]{article}
\usepackage[hyperref]{acl2020}
\usepackage{times}
\usepackage{latexsym}

\usepackage{microtype}

\usepackage{makecell}
\usepackage{multirow}
\usepackage{amsmath}
\usepackage{amssymb}
\usepackage{enumitem}
\usepackage{color}
\usepackage{graphicx}

\aclfinalcopy 

\title{An Empirical Investigation of Pre-Trained Transformer Language Models for Open-Domain Dialogue Generation}

\author{Piji Li\\
Tencent AI Lab\\
pijili@tencent.com}

\date{}

\begin{document}
\maketitle

\begin{abstract}
	We present an empirical investigation of pre-trained Transformer-based auto-regressive language models for the task of open-domain dialogue generation.  Training paradigm of pre-training and fine-tuning is employed to conduct the parameter learning. Corpora of News and Wikipedia in Chinese and English are collected for the pre-training stage respectively.  Dialogue context and response are concatenated into a single sequence utilized as the input of the models during the fine-tuning stage. A weighted joint prediction paradigm for both context and response is designed to evaluate the performance of models with or without the loss term for context prediction. Various of decoding strategies such as greedy search, beam search, top-k sampling, etc. are employed to conduct the response text generation.  
	Extensive experiments are conducted on the typical single-turn and multi-turn dialogue corpora such as Weibo, Douban, Reddit, DailyDialog, and Persona-Chat. Detailed numbers of automatic evaluation metrics on relevance and diversity of the generated results for the languages models as well as the baseline approaches are reported.\footnote{Code and data available at http://github.com/lipiji/Guyu.}

\end{abstract}

\section{Introduction}

Open-domain dialogue generation is an increasingly important task in the area of natural language processing and has received considerable critical attention across a number of researchers in recent years \cite{gao2019neural}.  Inspired by the frameworks utilized in the the task of neural machine translation \cite{bahdanau2015neural,cho2014learning}, the pioneer researchers formulate the dialogue generation task as a sequence-to-sequence (Seq2Seq) translation problem \cite{shang2015neural,vinyals2015neural,sordoni2015neural}.  Encoder is designed for dialogue context modeling, and decoder is constructed to conduct the target response prediction. Long Short-Term Memory (LSTM) \cite{hochreiter1997long}, Gated Recurrent Unit (GRU) \cite{cho2014learning}, and Transformer Self-Attention Network (SAN) \cite{vaswani2017attention}  are employed as the backbone of various encoders and decoders. For training, Maximum Likelihood Estimation (MLE) is usually as the parameter optimization approach. For decoding, beam search algorithm \cite{koehn2004pharaoh} is the preference to decode and generate the target response.

Tremendous progress has been achieved following those pioneer works. Simultaneously, on one side, researchers propose some strategies to address the typical issues observed from the generated results, and on the other side, amount of new directions and tasks have been explored broadly \cite{gao2019neural}. Specifically, one well-know typical phenomenon is that the generated responses are usually generic and boring. Quiet a few results are neither relevant to the context nor informative and diverse. This is caused by various of reasons, such as the MLE optimization objective, semantic or posterior collapse of the input context representations, one-to-many mapping issue (one single query can match several different but corresponding responses), etc. To address this issue, lots of model structures, training strategies, and operation tricks have been constructed and proposed, such as the Maximum Mutual Information (MMI) optimization objective  \cite{li2016diversity}, adversarial training \cite{li2017adversarial,xu2017neural,zhang2018generating} or reinforcement learning \cite{li2016deep,cuayahuitl2017simpleds,dhingra2017towards} tricks, continuous or discrete latent semantic modeling approaches \cite{zhao2017learning,gu2018dialogwae,gao2019generating,qiu2019training,gao2019discrete}, hierarchical encoder-decoder frameworks (HRED) for multi-turn dialogue generation \cite{serban2016building,serban2017hierarchical,park2018hierarchical}, etc.
Meanwhile, expanded from the general dialogue generation problem, more interesting and challenging tasks have been explored.
For instance, persona context \cite{li2016persona,zhang2018personalizing,mazare2018training,chan2019modeling,madotto2019personalizing,zheng2019pre,xu2020neural} and emotional information \cite{zhou2018emotional,zhou2018design,zhou2018emotional,huang2018automatic,rashkin2019towards,li2019empgan} are incorporated into the basic framework to improve the anthropomorphic characteristic. External knowledge such as the Commonsense Knowledge \cite{liu2004conceptnet} is also introduced to conduct the knowledge-grounded dialogue generation \cite{ghazvininejad2018knowledge,young2018augmenting,zhou2018commonsense,dinan2018wizard}. The problem of sentence function controlled dialogue generation has also been investigated \cite{ke2018generating,bi2019fine}.

Recently, pre-trained transformer-based language models such as BERT \cite{devlin2019bert} and GPT-2 \cite{radford2019language,radford2019better} have achieved tremendous success in various tasks of natural language processing， such as text classification, text matching, question answering, summarization \cite{dong2019unified}, storytelling \cite{see2019massively}, etc. Researchers are also trying to construct the transformer-based language models for the task of dialogue generation. Typically,  \citet{olabiyi2019multi} conduct some experimental analysis about the performance of transformer-based auto-regressive language models on dialogue generation. \citet{li2019relevance} propose a relevance-promoting language model by incorporating a topic inference component into the transformer language model to conduct diverse and informative response generation. \citet{zhang2019dialogpt} introduce a pre-training stage and MMI \cite{li2016diversity} is employed for results re-ranking and selection. \citet{adiwardana2020towards} release a chat-bot named Meena which is constructed based on Evolved Transformer (ET) \cite{so2019evolved} with 2.6B parameters and is trained on a large-scale multi-turn open domain dialogue corpus with 867M context-response pairs (341G text data with 40B words). A new human evaluation metric called Sensibleness and Specificity Average (SSA) is proposed and experimental results demonstrate that perplexity highly correlates with SSA.

In this paper, we present an empirical study of pre-trained transformer-based auto-regressive language models for the task of open-domain dialogue generation. The purpose is to construct a vividly sense and understanding of the performance for the (pre-trained) transformer-based auto-regressive language models on the tasks of open-domain dialogue generation. 
Training paradigm of pre-training and fine-tuning is employed to conduct the parameter learning. Corpora of News and Wikipedia in Chinese and English are collected for the pre-training stage respectively.  Dialogue context and response are concatenated into a single sequence which is utilized as the input of the language model during the fine-tuning stage. A weighted joint predicting paradigm for both context and response is designed to evaluate the performance of models  with or without the loss term for the context prediction results. Various of decoding strategies such as greedy search, beam search, top-k sampling, etc. are employed to conduct the response text prediction.  
Extensive experiments are conducted on the typical single-turn and multi-turn dialogue corpora such as Weibo \cite{shang2015neural,gao2019generating}, Douban \cite{pan2019improving}, Reddit \cite{zhou2018commonsense}, DailyDialog \cite{li2017dailydialog}, and Persona-Chat \cite{zhang2018personalizing}. Detailed numbers of automatic evaluation metrics on relevance (BLEU) and diversity (Distinct n-gram ratios) of the generated results  are reported.

In the following sections, we will first introduce the model architecture (Section \ref{sec:model}), and then the experimental settings (Section \ref{sec:settings}) and results discussions (Section~\ref{sec:results}). Finally, we will provide sufficient examples to demonstrate the performance (Section~\ref{sec:cases}) and following the conclusions as well as the future plans (Section~\ref{sec:cons}).

\section{Model Architecture}
\label{sec:model}

\begin{figure*}[!t]
	\centering
	\includegraphics[width=1.9\columnwidth]{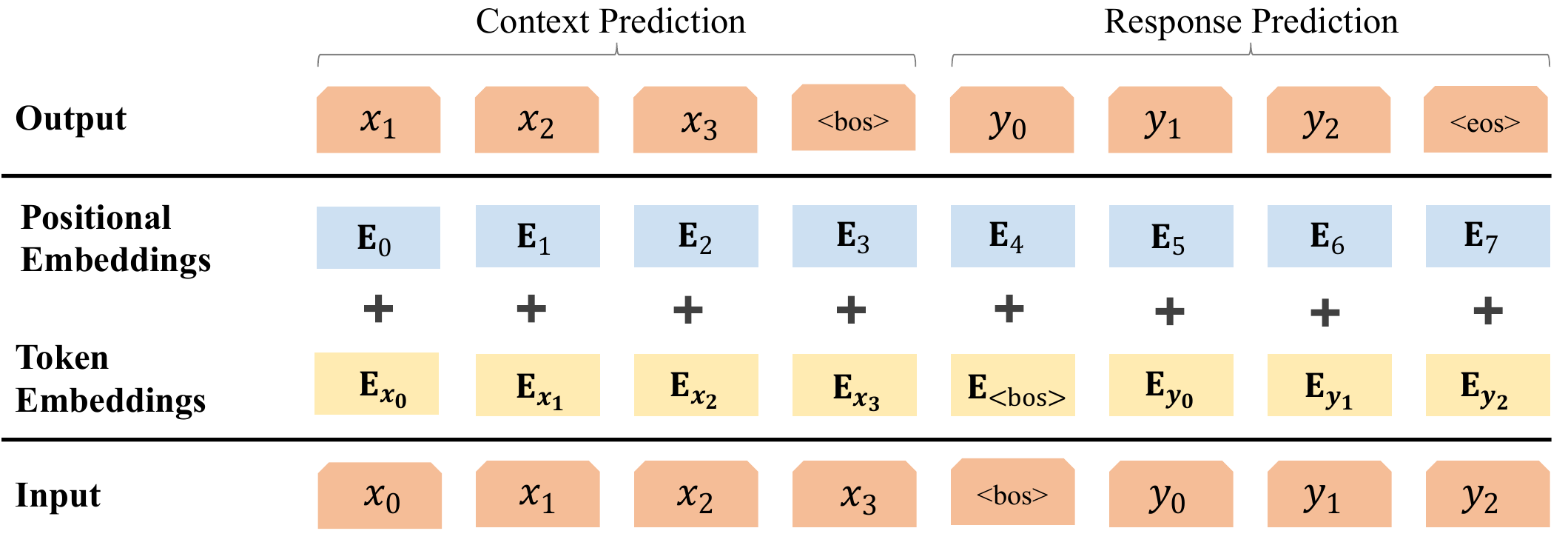}
	\caption{\label{fig:framework}
		Transformer-based auto-regressive language model for dialogue generation.
	}
\end{figure*}

As shown in Figure~\ref{fig:framework}, the core model architecture investigated in this work is a GPT-2 \cite{radford2019better} like transformer-based auto-regressive language model. We use $X = (x_0,x_1, x_2, \dots, x_m)$ and $Y = (y_0,y_1, y_2, \dots, y_n)$ to denote the sequence of input context and target response respectively. $m$ and $n$ represent the length of context and response,  respectively. It should be noted that we regard the input context as a whole sequence, either single-turn query or multi-turn historical context utterances.  Considering that the basic framework is a auto-regressive language model, therefore, we concatenate the context sequence $X$ and the response sequence $Y$ and obtain a whole sequence $I = (X, Y)$ as the input. For convenience, we assume that the  separator token $<bos>$ between $X$ and $Y$ is the last token $x_m \in X$, then we can denote the final sequence length as $T = |I| = m + n$.

Masking multi-head self-attention is utilized as the core technical operation to conduct representation learning:
\begin{equation}
\begin{split}
\mathrm{\bf H}^{1}_{t} &= \textsc{Ln}\left(\textsc{Ffn} (\mathrm{\bf H}^{1}_{t}) +\mathrm{\bf H}^1_t \right) \\
\mathrm{\bf H}^{1}_{t} &= \textsc{Ln}\left(\textsc{Slf-Att} (\mathrm{\bf Q}^{0}_{t}, \mathrm{\bf K}^{0}_{\leq t}, \mathrm{\bf V}^{0}_{\leq t}) +\mathrm{\bf H}^0_t \right) \\
\mathrm{\bf Q}^{0} &=  \mathrm{\bf H}^{0} \mathrm{\bf W}^{Q} \\
\mathrm{\bf K}^{0}, \mathrm{\bf V}^{0} &= \mathrm{\bf H}^{0}\mathrm{\bf W}^{K}, \mathrm{\bf H}^{0} \mathrm{\bf W}^{V}
\end{split}
\label{eql:formant_c1}
\end{equation}
where \textsc{Slf-Att}($\cdot$), \textsc{Ln}($\cdot$), and \textsc{Ffn}($\cdot$) represent self-attention mechanism, layer normalization, and feed-forward network respectively \cite{vaswani2017attention}. Note that we only use the states whose indices $\leq t$ as the attention context.

\noindent \textbf{Training}: During training, we employ Maximum Likelihood Estimation (MLE) to conduct the parameter learning and treat negative log-likelihood (NLL) over the whole target response sequence as the loss function:
\begin{equation}
\begin{split}
\mathcal{L}_{\mathrm{nll}}^{y} = -\sum^{n}_{t=1} \log P(\mathrm{y}_t|\mathrm{y}_{<t},X)
\end{split}
\label{eql:nll_y}
\end{equation}

Intuitively, since we have concatenated $X$ and $Y$, then we can obtain another term of prediction loss, i.e., the loss term for predicting the dialogue context $X$.  
\begin{equation}
\begin{split}
\mathcal{L}_{\mathrm{nll}}^{x} = -\sum^{m}_{t=1} \log P(\mathrm{x}_t|\mathrm{x}_{<t})
\end{split}
\label{eql:nll_x}
\end{equation} 

Considering that context and response are monolingual pairs, then introduce the context prediction loss may benefit the language modeling performance, since context texts contribute the other half of corpus to teach the model how to handle a language. This assumption is also mentioned in the previous works of \citet{olabiyi2019multi} and \citet{li2019relevance}. Therefore, a weighted joint predicting paradigm for both context and response is constructed. By tuning the weight, we can also  evaluate the performance of models  with or without the loss term for the context prediction results.
\begin{equation}
\begin{split}
\mathcal{L}_{\mathrm{nll}} = \mathcal{L}_{\mathrm{nll}}^{y}  + \lambda \times \mathcal{L}_{\mathrm{nll}}^{x} 
\end{split}
\label{eql:nll}
\end{equation}

\noindent \textbf{Decoding}: In the inference stage, various of decoding strategies such as greedy search, beam search \cite{koehn2004pharaoh}, truncated top-$k$ sampling  \cite{fan2018hierarchical,radford2019language}, and Nucleus Sampling (top-$p$ sampling) \cite{holtzman2019curious} are employed to conduct the response text generation. Hyper-parameter such as $k$ is tuned to trade-off the performance of relevance and diversity.

\section{Experimental Settings}
\label{sec:settings}

\subsection{Datasets}

\noindent \textbf{Pre-training Corpus}: To investigate the performance of models in different language scenarios, we conduct the pre-training procedure on two large scale corpora in Chinese and English respectively. 
\begin{itemize}[leftmargin=*,topsep=0pt]
	\setlength\itemsep{-0.5em}
	\item \textbf{Chinese}: We collect Chinese Wikipedia\footnote{https://dumps.wikimedia.org/zhwiki/latest/} (1.7B Characters) and a merged Chinese News corpus (9.2B Characters) from the Internet. We did not conduct the word segmenting operations on the Chinese datasets, which means that we just use the characters to build the vocabulary, and the size is 27,681.
	\item \textbf{English}: We employ English Wikipedia\footnote{https://dumps.wikimedia.org/enwiki/latest/} (2.4B words) and BooksCorpus (980M words) \cite{zhu2015aligning} to conduct the pre-training. Byte Pair Encodings (BPE) \cite{sennrich2016neural} is conducted on this corpus as the tokenizer and the vocabulary size is 32,000.
\end{itemize}

\noindent \textbf{Fine-tuning Dialogue Corpus}
Five typical dialogue corpora are collected to train the dialogue generation models: Weibo \cite{shang2015neural,gao2019generating}, Douban \cite{pan2019improving}, Reddit \cite{zhou2018commonsense}, DailyDialog \cite{li2017dailydialog}, and Persona-Chat \cite{zhang2018personalizing}. These corpora are in different languages with single-turn or multi-turn historical dialogue context. Moreover, some  pre-defined external information are also provided in the dataset of Persona-Chat besides the pure dialogue utterances.
\begin{itemize}[leftmargin=*,topsep=0pt]
	\setlength\itemsep{-0.5em}
	\item \textbf{Weibo}:  A single-turn open-domain Chinese dialogue dataset, which is originally collected and released by \citet{shang2015neural} from a Chinese social media platform Weibo\footnote{https://www.weibo.com/}. Here, we use a refined version released by \citet{gao2019generating} \footnote{https://ai.tencent.com/ailab/nlp/dialogue/\#datasets \label{ft:ailab-data}}. The number of samples for training, validation, and testing are 400M, 19357, and 3200 respectively. Character-based vocabulary size is 10231.
	\item \textbf{Douban}:  A multi-turn open-domain Chinese dialogue corpus collected from Douban group\footnote{https://www.douban.com/group}, a well-known Chinese online community which is also a common data source for dialogue systems \cite{wu2017sequential}. Here, we utilize a version called Restoration-200K dataset released by \citet{pan2019improving} \textsuperscript{\ref{ft:ailab-data}}. There are 193769 samples for training, 5095 for validation, and 5104 for testing. Vocabulary size is 5800.
	\item \textbf{Reddit}: A single-turn open-domain English corpus. We use the version collected and processed by \citet{zhou2018commonsense} \footnote{http://coai.cs.tsinghua.edu.cn/file/commonsense\_conversation\\\_dataset.tar.gz}. The original version is designed for commonsense dialogue generation, and here we only keep the dialogue utterances and filter the commonsense knowledge annotations. There are 300M samples for training, 10000 for validation, and 20000 for testing. BPE vocabulary size is 14820.
	\item \textbf{DailyDialog}: A multi-turn open-domain English corpus released by \citet{li2017dailydialog} \footnote{http://yanran.li/dailydialog.html}. There are 11118 for training, 1000 for validation, and 1000 for testing. BPE vocabulary size is 12244.
	\item \textbf{Persona-Chat}:  Another multi-turn open-domain English corpus, which contains 8939 samples for training, 1000 for validation, and 968 for testing after filtering \cite{zhang2018personalizing}. Moreover, this dataset provides 3-5 sentences to describe the persona information for each speaker (e.g. ``\textit{I like reading.}'' or ``\textit{I am a nurse.}'', etc). In our tasks, we concatenate the sequence of persona sentences before the dialogue context as another kind of condition to control the response generation. BPE vocabulary size is 11362.
\end{itemize}

\begin{table*}[!t]
	\centering
	\begin{tabular}{l|c|c|c|c|c|c}
		\Xhline{3\arrayrulewidth}
		Corpus & Language & Type & \#Train & \#Dev & \#Test & \#Vocab \\
		\hline
		\hline
		Weibo & Chinese & Single-Turn & 4,244,093 & 19,357 &  3,200 & 10,231  \\
		\hline
		Douban & Chinese & Multi-Turn & 193,769 & 5,095 & 5,104 & 5,800    \\
		\hline
		Reddit & English & Single-Turn & 3,384,185 & 10,000 & 20,000 & 14,820    \\
		\hline
		DailyDialog & English & Multi-Turn & 11,118 & 1,000 & 1,000 & 12,244    \\
		\hline
		Persona-Chat & English & Multi-Turn & 8,939 & 1,000 & 968 & 11,362    \\
		\Xhline{3\arrayrulewidth}
	\end{tabular}
	\caption{Statistics of the dialogue datasets.}
	\label{tab:dialogue-corpora}
\end{table*}

The specific statistic numbers of the dialogue corpus are depicted in Table~\ref{tab:dialogue-corpora}. We can observe that Weibo and Reddit can be treated as large-scale corpora, and Douban is a midle-size dataset. Comparatively, DailyDialog and Persona-Chat are in small-scale.

\subsection{Metrics}
As did in the previous works, we employ BLEU \footnote{https://github.com/tensorflow/nmt/blob/master/nmt/scripts/bleu.py} and Distinct n-gram ratios as the automatic evaluation metrics. For BLEU, we provide the values of BLEU 1-4 as well as a weighted summation of them. For Distinct, we calculate the uni-gram and bi-gram distinct rations and both macro-average (\textsc{MaDist}, average on instance level) and micro-average (\textsc{MiDist}, average on corpus level) values are reported. Moreover, we also provide the average-length of the generated response for reference. More automatic metrics such as PPL, Embedding Matching, METEOR, etc., as well as the human evaluation performance will be completed in the future.

\subsection{Comparison Methods}

\begin{itemize}[leftmargin=*]
	\setlength\itemsep{0pt}
	\item \textbf{Seq2Seq}: The vanilla Transformer Sequence-to-Sequence model used for machine translation where the network components are identical to those in~\cite{vaswani2017attention}. 
	\item \textbf{CVAE}~\cite{zhao2017learning}: Variational Auto-Encoder (VAE) \cite{kingma2013auto} is introduced into the LSTM-based seq2seq framework for context semantic modeling and response generation.
	\item \textbf{MMI}~\cite{li2016diversity}: LSTM-based seq2seq framework with Maximum Mutual Information (MMI) objective in decoding . In this paper, we set the number of responses for re-ranking as 50.
	\item \textbf{HRED}~\cite{serban2016building}:  A hierarchical RNN-based encoder-decoder constructed for the tasks of multi-turn dialogue generation.   
	\item \textbf{VHRED}~\cite{serban2017hierarchical}: Variational Auto-Encoder (VAE) \cite{kingma2013auto} is incorporated into the HRED model for better semantic modeling for each dialogue utterance.
	\item \textbf{VHCR}~\cite{park2018hierarchical}: An extension of VHRED, where a hierarchical latent variable model with the utterance drop regularization is proposed to enhance the global semantic modeling of the dialogue context.
	\item \textbf{LM-12} and \textbf{LM-24}: Transformer-based auto-regressive language model with layers of 12 and 24 respectively. \textit{No pre-training stage for these models. We only conduct the training purely on the dialogue corpus.}  The suffix of $greedy$, $bm(b)$, $tk(k)$, $tp(p)$ denote the different decoding strategies, i.e.,  greedy search, beam search \cite{koehn2004pharaoh}, truncated top-$k$ sampling  \cite{fan2018hierarchical,radford2019language}, and Nucleus Sampling (top-$p$ sampling) \cite{holtzman2019curious}. $b$, $k$, and $p$ are three hyper-parameters.
	\item \textbf{*-$\lambda$} (Eq.~\ref{eql:nll}): $\lambda$ is the weight of loss term for dialogue context prediction. $\lambda_{0.5}$ and $\lambda_1$ denote that $\lambda$ is 0.5 and 1 respectively.  The models without the term $\lambda$ means that $\lambda = 0$, i.e., only conduct the response prediction given the dialogue context.
	\item \textbf{BigLM-12} and \textbf{BigLM-24}:  The language models with both the pre-training and fine-tuning procedures. 
\end{itemize}

\subsection{Settings}
The length limitation of dialogue context and response for single-turn dialogue dataset are both 64, then the maximum of the whole sequence length is 128. For the multi-turn tasks, the length limitation is 300 and 64 for context and response respectively.

LM-12 and BigLM-12 are the typical GPT-2 model with 117M parameters (768 dimensions, 12 layers, 12 heads). LM-24 and BigLM-24 only expand the number of layers from 12 to 24, and the typical 345M models (1024 dimensions, 24 layers, 16 heads) will be evaluated and provided later.

For LM and BigLM models, Adam \cite{kingma2014adam} optimization method with Noam learning-rate decay strategy is employed to conduct the parameter learning. Validation datasets  are used to select the appropriate models. For Seq2Seq, Adam with learning rate $1e$-$5$ is utilized for training.

\section{Results and Discussions}
\label{sec:results}

Table~\ref{tab:result-weibo}$\sim$Table~\ref{tab:result-persona} depict the detailed evaluation numbers for all the models on datasets of Weibo, Douban, Reddit, DailyDialog, and Persona-Chat respectively.  We will conduct the analysis and discussions according to the following aspects \footnote{There may be some shakes behind the numbers due to the random sampling issue. More robust evaluation approaches need to be designed. }:

\begin{itemize}[leftmargin=*]
	\setlength\itemsep{0pt}
	\item Compared with the vanilla transformer-based Seq2Seq framework, auto-regressive language models can indeed improve the performance on most of the datasets. This phenomenon is more clear on the multi-turn datasets of Douban, DailyDialog, and Persona-Chat. On those datasets, metric values on BLEU or BLEU-n are much better than Seq2Seq. On Weibo, LM models also obtain comparable performance. However, on Reddit dataset, Seq2Seq achieves the best performance. 
	\item Obviously, pre-training can benefit the performance on relevance. From the results we can observe that the models with pre-training stage, i.e., BigLM-12 and BigLM-24, obtain better BLEU scores than LM-12 and LM-24 on all the five datasets.
	\item Models with context prediction loss term (*-$\lambda$) can also improve the performance, especially on multi-turn dialogue scenario. From the results on Douban, DailyDialog, and Persona-Chat we see that LM-12-$\lambda_1$ models are usually better than the original model LM-12 on metric of BLEU. Also, LM-12-$\lambda_1$ is generally better than LM-12-$\lambda_{0.5}$. However, from datasets of Weibo and Reddit, we cannot get this kind of observation and clue. The reason may be that long context can provide more useful information than the single-turn query with only one sentence. 
	\item For top-$k$ sampling decoding methods, $k$ is a sensitive parameter to trade-off the relevance and diversity. Large $k$ introduces better diversity but decreases the relevance. Thus the tuning of $k$ is also an crucial issue.
	\item Parameter $k$ can affect the length of the generated response. Large $k$ causes longer responses. The reason is that a large sampling pool will decrease the opportunity of the special token $<$eos$>$.
	\item Nucleus Sampling (top-$p$ sampling) sometimes obtain comparable or better performance than top-$k$ sampling. But the improvements are not strong and  steady.
	\item CVAE and MMI are still the strong baseline methods and always produce better performance on BLEU and Distinct metrics, which demonstrates that latent variable modeling and MMI strategy can further improve the quality of the generated results.  We believe that incorporating those strategies into the LM framework may further enhance the performance significantly.
\end{itemize}

\begin{table*}[]
    \centering
    \resizebox{1.9\columnwidth}{!}{
    \begin{tabular}{l|ccccc|cccc|c}
    \Xhline{3\arrayrulewidth}
     \multirow{2}{*}{\textbf{Model}} & \multicolumn{5}{c|}{\textbf{Relevance}} & \multicolumn{4}{c|}{\textbf{Diversity}} & \multirow{2}{*}{\textbf{Length}} \\
     \cline{2-10}
    & \textsc{Bleu} & \textsc{B-1} & \textsc{B-2} & \textsc{B-3} & \textsc{B-4} & \textsc{MaDist-1} & \textsc{MaDist-2} & \textsc{MiDist-1} & \textsc{MiDist-2} \\ \hline \hline
    Seq2Seq & 1.31 & 37.08 & 10.29 & 3.76 & 1.71 & 71.73 & 75.53 & 7.70 & 26.63 & 13.88 \\
    CVAE & 4.08 & 42.12 & 15.90 & 8.13 & 5.31 & 73.61 & 78.09 & 9.59 & 34.64 & 14.79 \\ 
    MMI & 3.25 & 43.98 & 13.94 & 6.56 & 3.88 & 82.00 & 87.06 & 10.69 & 38.25 & 11.51\\
    \hline \hline
    Seq2Seq-greedy& 1.09 & 41.62 & 8.99 & 2.85 & 0.96 & 77.20 & 83.81 & 7.56 & 29.39 & 14.31\\
    Seq2Seq-bm(5) & 1.31 & 37.08 & 10.29 & 3.76 & 1.71 & 71.73 & 75.53 & 7.70 & 26.63 & 13.88\\
    Seq2Seq-tk(5) & 0.72 & 41.82 & 7.82 & 1.92 & 0.53 & 86.10 & 94.50 & 9.71 & 44.54 & 13.62\\
    Seq2Seq-tk(10) & 0.59 & 41.21 & 7.05 & 1.75 & 0.54 & 87.22 & 95.94 & 10.01 & 50.17 & 13.90\\
    Seq2Seq-tk(20) & 0.63 & 39.06 & 6.07 & 1.50 & 0.42 & 87.90 & 96.27 & 9.78 & 52.62 & 14.39\\
    Seq2Seq-tk(50) & 0.22 & 35.72 & 4.73 & 0.74 & 0.17 & 89.45 & 97.59 & 10.45 & 58.09 & 14.87\\
    Seq2Seq-tk(500) & 0.26 & 30.74 & 3.69 & 0.65 & 0.17 & 91.28 & 98.14 & 12.58 & 67.98 & 16.09\\
    Seq2Seq-tp(0.9) & 0.59 & 39.10 & 6.43 & 1.52 & 0.50 & 88.35 & 96.56 & 9.99 & 53.01 & 14.43\\
    \hline
    LM-12-greedy & 1.32 & 42.16 & 8.41 & 2.74 & 1.40 & 81.43 & 86.94 & 6.44 & 21.24 & 13.50\\
    LM-12-bm(5) & 2.56 & 35.04 & 11.45 & 5.39 & 3.29 & 63.66 & 66.96 & 4.70 & 14.55 & 21.19\\
    LM-12-tk(5) & 0.79 & 41.27 & 6.72 & 1.65 & 0.65 & 85.77 & 94.49 & 7.76 & 35.55 & 13.64\\
    LM-12-tk(10) & 0.65 & 40.06 & 6.17 & 1.40 & 0.51 & 87.10 & 95.54 & 7.98 & 40.21 & 14.17\\
    LM-12-tk(20) & 0.35 & 37.98 & 5.14 & 1.03 & 0.25 & 88.63 & 96.79 & 8.91 & 46.42 & 14.46\\
    LM-12-tk(50) &0.31 & 36.53 & 4.67 & 0.82 & 0.24 & 89.85 & 97.53 & 9.24 & 52.24 & 15.03\\
    LM-12-tk(500) & 0.25 & 29.81 & 3.07 & 0.54 & 0.19 & 91.71 & 98.43 & 12.11 & 67.41 & 15.77\\
    LM-12-tp(0.9)& 0.47 & 38.58 & 5.34 & 1.04 & 0.28 & 88.75 & 96.85 & 8.63 & 45.48 & 14.42\\
    \hline
    LM-12-greedy-$\lambda_1$ & 0.96 & 43.20 & 8.14 & 2.20 & 0.90 & 80.93 & 87.84 & 5.68 & 18.41 & 13.81\\
    LM-12-bm(5)-$\lambda_1$ & 1.86 & 30.34 & 8.90 & 4.04 & 2.46 & 55.86 & 59.11 & 3.30 & 10.00 & 24.33\\
    LM-12-tk(5)-$\lambda_1$ & 0.50 & 41.79 & 6.50 & 1.42 & 0.44 & 85.91 & 94.81 & 7.09 & 32.53 & 13.76\\
    LM-12-tk(10)-$\lambda_1$ & 0.53 & 40.33 & 6.02 & 1.28 & 0.37 & 87.25 & 96.04 & 7.73 & 39.05 & 14.15\\
    LM-12-tk(20)-$\lambda_1$ & 0.47 & 37.51 & 4.82 & 1.17 & 0.41 & 88.66 & 96.85 & 8.22 & 44.89 & 14.69\\
    LM-12-tk(50)-$\lambda_1$ & 0.31 & 35.78 & 4.38 & 0.79 & 0.23 & 89.78 & 97.40 & 9.12 & 52.50 & 15.05\\
    LM-12-tk(500)-$\lambda_1$ & 0.21 & 29.29 & 3.15 & 0.56 & 0.13 & 91.57 & 98.14 & 11.61 & 65.94 & 15.88\\
    LM-12-tp(0.9)-$\lambda_1$ & 0.40 & 38.56 & 5.35 & 1.15 & 0.37 & 88.41 & 96.92 & 8.04 & 43.98 & 14.66\\
    \hline
    LM-12-greedy-$\lambda_{0.5}$& 1.03 & 41.00 & 7.74 & 2.42 & 1.02 & 80.79 & 86.45 & 5.59 & 18.60 & 13.42\\
    LM-12-bm(5)-$\lambda_{0.5}$ & 1.60 & 31.58 & 8.80 & 3.69 & 2.00 & 59.20 & 62.23 & 3.72 & 11.56 & 22.87\\
    LM-12-tk(5)-$\lambda_{0.5}$ & 0.59 & 41.71 & 6.19 & 1.47 & 0.50 & 85.21 & 94.10 & 7.15 & 33.68 & 13.94\\
    LM-12-tk(10)-$\lambda_{0.5}$ & 0.36 & 39.85 & 5.50 & 1.05 & 0.25 & 87.76 & 96.03 & 7.57 & 38.20 & 13.98\\
    LM-12-tk(20)-$\lambda_{0.5}$ & 0.38 & 37.75 & 5.14 & 1.03 & 0.32 & 88.43 & 96.79 & 8.14 & 45.28 & 14.43\\
    LM-12-tk(50)-$\lambda_{0.5}$ & 0.34 & 35.01 & 4.27 & 0.89 & 0.32 & 89.27 & 96.93 & 8.95 & 51.60 & 14.68\\
    LM-12-tk(500)-$\lambda_{0.5}$ & 0.29 & 29.37 & 3.21 & 0.59 & 0.15 & 92.02 & 98.47 & 11.40 & 66.33 & 16.11\\
    LM-12-tp(0.9)-$\lambda_{0.5}$ & 0.26 & 37.65 & 4.87 & 0.89 & 0.18 & 88.52 & 96.83 & 8.37 & 45.25 & 14.27\\
    \hline
    LM-24-greedy& 1.12 & 39.21 & 7.52 & 2.42 & 1.19 & 75.69 & 81.98 & 5.21 & 17.79 & 16.02\\
    LM-24-bm(5) & 2.02 & 34.18 & 9.80 & 4.22 & 2.52 & 64.63 & 68.30 & 4.73 & 15.61 & 19.86\\
    LM-24-tk(5) & 0.90 & 41.04 & 6.82 & 1.76 & 0.70 & 84.90 & 94.00 & 7.00 & 32.57 & 14.64\\
    LM-24-tk(10) & 0.48 & 39.68 & 5.60 & 1.16 & 0.38 & 86.18 & 95.42 & 7.43 & 37.57 & 14.99\\
    LM-24-tk(20) & 0.60 & 37.38 & 4.88 & 1.09 & 0.48 & 87.32 & 95.93 & 7.77 & 43.03 & 15.07\\
    LM-24-tk(50) & 0.33 & 34.39 & 4.03 & 0.77 & 0.24 & 89.10 & 97.13 & 8.81 & 51.42 & 15.67\\
    LM-24-tk(500) & 0.21 & 28.20 & 2.92 & 0.51 & 0.24 & 91.07 & 97.80 & 11.12 & 65.57 & 16.83\\
    LM-24-tp(0.9) & 0.45 & 37.17 & 5.25 & 1.16 & 0.39 & 86.64 & 95.85 & 7.86 & 43.87 & 15.56\\
    \hline \hline
    BigLM-12-greedy & 2.07 & 41.92 & 9.65 & 3.48 & 1.99 & 79.88 & 86.06 & 7.05 & 24.32 & 14.62 \\
    BigLM-12-bm(5)  & 4.00 & 38.96 & 14.77 & 7.70 & 5.14 & 72.61 & 75.97 & 7.50 & 25.97 & 18.62\\
    BigLM-12-tk(5) & 1.01 & 43.18 & 8.08 & 2.31 & 1.01 & 85.81 & 93.47 & 8.74 & 39.23 & 13.78\\
    BigLM-12-tk(10) & 0.76 & 40.69 & 6.99 & 1.87 & 0.69 & 87.61 & 95.78 & 9.23 & 44.20 & 14.34\\
    BigLM-12-tk(20) & 0.95 & 39.28 & 6.25 & 1.78 & 0.76 & 88.62 & 96.76 & 9.02 & 48.06 & 14.54\\
    BigLM-12-tk(50) & 0.53 & 35.56 & 4.85 & 1.07 & 0.46 & 90.18 & 97.32 & 9.84 & 55.06 & 14.95\\
    BigLM-12-tk(500) & 0.18 & 29.49 & 3.34 & 0.45 & 0.11 & 91.09 & 97.68 & 12.26 & 67.49 & 15.56\\
    BigLM-12-tp(0.9) & 0.95 & 38.78 & 6.24 & 1.80 & 0.80 & 88.53 & 96.45 & 9.27 & 48.38 & 14.13\\
    \hline
    BigLM-24-greedy & 2.32 & 43.38 & 10.68 & 4.13 & 2.39 & 81.06 & 86.31 & 7.17 & 24.72 & 14.22\\
    BigLM-24-bm(5) & 3.97 & 38.93 & 14.72 & 7.59 & 5.07 & 70.56 & 73.84 & 7.24 & 24.87 & 18.95 \\
    BigLM-24-tk(5) & 1.15 & 43.64 & 8.45 & 2.55 & 1.17 & 85.54 & 93.67 & 8.44 & 38.07 & 13.75\\
    BigLM-24-tk(10) & 1.22 & 41.23 & 8.01 & 2.49 & 1.22 & 86.71 & 95.42 & 8.81 & 43.83 & 14.45\\
    BigLM-24-tk(20) & 0.75 & 37.53 & 5.79 & 1.52 & 0.83 & 88.16 & 96.06 & 9.59 & 48.65 & 14.25\\
    BigLM-24-tk(50) & 0.57 & 36.36 & 5.12 & 1.23 & 0.54 & 89.39 & 97.24 & 9.78 & 53.64 & 14.61\\
    BigLM-24-tk(500) & 0.39 & 29.51 & 3.49 & 0.85 & 0.39 & 90.89 & 97.63 & 12.22 & 67.21 & 15.82\\
    BigLM-24-tp(0.9) & 0.93 & 39.00 & 5.97 & 1.73 & 0.76 & 88.57 & 96.48 & 9.21 & 48.53 & 14.31\\
    \Xhline{3\arrayrulewidth}
    \end{tabular}}
    \caption{Experimental results on Weibo dataset.}
    \label{tab:result-weibo}
\end{table*}

\begin{table*}[]
    \centering
    \resizebox{1.9\columnwidth}{!}{
    \begin{tabular}{l|ccccc|cccc|c}
    \Xhline{3\arrayrulewidth}
     \multirow{2}{*}{\textbf{Model}} & \multicolumn{5}{c|}{\textbf{Relevance}} & \multicolumn{4}{c|}{\textbf{Diversity}} & \multirow{2}{*}{\textbf{Length}} \\ \cline{2-10}
    & \textsc{Bleu} & \textsc{B-1} & \textsc{B-2} & \textsc{B-3} & \textsc{B-4} & \textsc{MaDist-1} & \textsc{MaDist-2} & \textsc{MiDist-1} & \textsc{MiDist-2} \\ \hline \hline
    Seq2Seq & 0.15 & 10.95 & 1.97 & 0.48 & 0.24 & 85.37 & 89.72 & 2.26 & 11.25 & 7.83\\
    CVAE & 0.09 & 15.14 & 4.84 & 0.64 & 0.13 & 92.23 & 93.12 & 5.25 & 22.46 & 3.93 \\ 
    MMI  & 0.06 & 9.83 & 1.91 & 0.35 & 0.09 & 89.95 & 94.52 & 5.46 & 37.55 & 5.20\\
    HRED & 0.06 & 13.13 & 2.67 & 0.36 & 0.07 & 92.08 & 93.02 & 3.09 & 11.72 & 4.24\\
    VHRED & 0.05 & 11.47 & 2.48 & 0.60 & 0.11 & 81.22 & 83.03 & 1.76 & 6.61 & 6.84 \\
    VHCR & 0.12 & 11.83 & 2.73 & 0.58 & 0.15 & 77.97 & 79.10 & 1.30 & 5.13 & 8.44 \\
    \hline \hline
    Seq2Seq-greedy & 0.12 & 11.74 & 1.64 & 0.35 & 0.15 & 88.17 & 92.14 & 2.35 & 15.98 & 9.49 \\
    Seq2Seq-bm(5) & 0.15 & 10.95 & 1.97 & 0.48 & 0.24 & 85.37 & 89.72 & 2.26 & 11.25 & 7.83\\
    Seq2Seq-tk(5) & 0.05 & 10.34 & 1.21 & 0.24 & 0.05 & 91.33 & 97.06 & 3.10 & 26.71 & 8.62\\
    Seq2Seq-tk(10) & 0.06 & 9.65 & 1.07 & 0.21 & 0.04 & 91.21 & 97.49 & 3.16 & 30.11 & 9.24\\
    Seq2Seq-tk(20) & 0.05 & 8.62 & 0.80 & 0.10 & 0.04 & 92.17 & 98.11 & 3.31 & 34.11 & 9.53\\
    Seq2Seq-tk(50) & 0.04 & 7.76 & 0.72 & 0.14 & 0.04 & 92.50 & 98.28 & 3.58 & 39.14 & 9.84\\
    Seq2Seq-tk(500) & 0.04 & 6.60 & 0.62 & 0.13 & 0.03 & 93.96 & 98.83 & 4.89 & 49.16 & 10.18\\
    Seq2Seq-tp(0.9) & 0.07 & 8.66 & 0.90 & 0.19 & 0.11 & 92.38 & 98.12 & 3.33 & 34.21 & 9.51\\
    \hline
    LM-12-greedy & 0.17 & 17.09 & 4.06 & 0.94 & 0.25 & 94.31 & 96.18 & 3.64 & 17.11 & 6.44\\
    LM-12-bm(5)  & 0.08 & 17.19 & 5.66 & 1.06 & 0.16 & 95.22 & 96.75 & 4.57 & 13.53 & 4.03\\
    LM-12-tk(5) & 0.13 & 13.67 & 2.53 & 0.53 & 0.14 & 92.65 & 97.31 & 3.75 & 25.50 & 6.80\\
    LM-12-tk(10) & 0.09 & 11.93 & 2.12 & 0.45 & 0.11 & 92.43 & 97.52 & 3.70 & 28.23 & 7.37\\
    LM-12-tk(20) & 0.09 & 10.79 & 1.67 & 0.32 & 0.09 & 92.41 & 97.77 & 3.62 & 31.76 & 7.76 \\
    LM-12-tk(50) & 0.05 & 9.42 & 1.55 & 0.27 & 0.08 & 93.00 & 98.14 & 3.80 & 36.99 & 8.49\\
    LM-12-tk(500) & 0.02 & 7.56 & 1.01 & 0.13 & 0.03 & 93.74 & 98.71 & 4.83 & 49.22 & 8.97\\
    LM-12-tp(0.9) & 0.06 & 10.93 & 1.90 & 0.41 & 0.12 & 92.83 & 97.97 & 3.73 & 32.15 & 7.79\\
    \hline
    LM-12-greedy-$\lambda_1$ & 0.18 & 17.09 & 3.96 & 1.07 & 0.26 & 94.67 & 96.77 & 3.91 & 18.43 & 6.01\\
    LM-12-bm(5)-$\lambda_1$  & 0.14 & 17.18 & 5.40 & 1.14 & 0.20 & 96.56 & 97.83 & 5.01 & 15.61 & 3.55\\
    LM-12-tk(5)-$\lambda_1$ & 0.07 & 13.43 & 2.43 & 0.55 & 0.10 & 92.77 & 97.37 & 3.78 & 25.83 & 6.89\\
    LM-12-tk(10)-$\lambda_1$ & 0.08 & 12.55 & 2.43 & 0.49 & 0.11 & 92.51 & 97.62 & 3.70 & 28.43 & 7.56\\
    LM-12-tk(20)-$\lambda_1$ & 0.11 & 10.88 & 1.83 & 0.42 & 0.11 & 92.28 & 97.88 & 3.72 & 31.77 & 8.05 \\
    LM-12-tk(50)-$\lambda_1$ & 0.06 & 9.39 & 1.30 & 0.21 & 0.06 & 92.46 & 98.12 & 3.72 & 36.92 & 8.74\\
    LM-12-tk(500)-$\lambda_1$ & 0.04 & 7.26 & 0.99 & 0.15 & 0.05 & 93.42 & 98.67 & 4.54 & 48.47 & 9.52\\
    LM-12-tp(0.9)-$\lambda_1$ & 0.07 & 10.91 & 1.67 & 0.32 & 0.06 & 92.38 & 97.89 & 3.62 & 31.66 & 8.13\\
    \hline
    LM-12-greedy-$\lambda_{0.5}$ & 0.14 & 16.87 & 4.22 & 1.06 & 0.25 & 93.76 & 95.88 & 3.66 & 17.14 & 6.49\\
    LM-12-bm(5)-$\lambda_{0.5}$  & 0.08 & 17.22 & 5.88 & 1.01 & 0.11 & 95.35 & 96.72 & 4.62 & 13.49 & 3.97\\
    LM-12-tk(5)-$\lambda_{0.5}$ & 0.12 & 13.59 & 2.59 & 0.46 & 0.11 & 92.69 & 97.36 & 3.82 & 26.14 & 6.87\\
    LM-12-tk(10)-$\lambda_{0.5}$ & 0.10 & 11.98 & 2.03 & 0.42 & 0.12 & 91.92 & 97.35 & 3.73 & 28.79 & 7.45\\
    LM-12-tk(20)-$\lambda_{0.5}$ & 0.09 & 10.87 & 1.82 & 0.43 & 0.11 & 92.45 & 97.86 & 3.84 & 32.53 & 7.82 \\
    LM-12-tk(50)-$\lambda_{0.5}$ & 0.09 & 9.63 & 1.52 & 0.25 & 0.08 & 92.53 & 98.02 & 3.85 & 37.64 & 8.36\\
    LM-12-tk(500)-$\lambda_{0.5}$ & 0.04 & 7.27 & 0.90 & 0.15 & 0.04 & 93.64 & 98.54 & 4.71 & 49.50 & 9.16\\
    LM-12-tp(0.9)-$\lambda_{0.5}$& 0.09 & 10.58 & 1.75 & 0.37 & 0.10 & 92.24 & 97.60 & 3.65 & 31.89 & 7.92\\
    \hline
    LM-24-greedy & 0.17 & 17.24 & 4.19 & 1.02 & 0.27 & 95.22 & 97.33 & 3.85 & 18.83 & 5.80\\
    LM-24-bm(5) & 0.07 & 16.68 & 5.50 & 0.94 & 0.11 & 96.64 & 98.02 & 4.99 & 15.37 & 3.40\\
    LM-24-tk(5) & 0.10 & 13.80 & 2.70 & 0.56 & 0.11 & 92.63 & 97.29 & 3.86 & 26.03 & 6.60\\
    LM-24-tk(10) & 0.10 & 12.14 & 2.03 & 0.37 & 0.08 & 92.56 & 97.51 & 3.79 & 28.76 & 7.09\\
    LM-24-tk(20) & 0.05 & 10.66 & 1.71 & 0.26 & 0.05 & 92.70 & 97.87 & 3.77 & 32.12 & 7.63\\
    LM-24-tk(50) & 0.09 & 9.63 & 1.65 & 0.40 & 0.11 & 92.99 & 98.12 & 3.87 & 37.65 & 8.15\\
    LM-24-tk(500) & 0.03 & 7.16 & 0.86 & 0.10 & 0.03 & 93.99 & 98.83 & 4.81 & 49.86 & 8.99\\
    LM-24-tp(0.9)  & 0.05 & 10.78 & 1.76 & 0.34 & 0.05 & 92.72 & 97.99 & 3.83 & 32.40 & 7.53\\
    \hline \hline
    BigLM-12-greedy  & 0.17 & 17.09 & 4.25 & 1.07 & 0.28 & 93.39 & 95.76 & 4.19 & 20.70 & 6.71\\
    BigLM-12-bm(5)  & 0.09 & 16.69 & 5.64 & 0.99 & 0.20 & 95.15 & 96.64 & 4.95 & 16.22 & 3.99\\
    BigLM-12-tk(5) & 0.09 & 13.75 & 2.76 & 0.59 & 0.12 & 91.99 & 96.65 & 4.28 & 28.47 & 6.93\\
    BigLM-12-tk(10) & 0.09 & 12.48 & 2.36 & 0.49 & 0.12 & 92.35 & 97.36 & 4.13 & 31.06 & 7.45\\
    BigLM-12-tk(20) & 0.10 & 10.98 & 1.89 & 0.46 & 0.13 & 92.53 & 97.59 & 4.15 & 34.23 & 7.83\\
    BigLM-12-tk(50) & 0.05 & 9.57 & 1.51 & 0.28 & 0.05 & 93.00 & 98.03 & 4.27 & 38.92 & 8.20\\
    BigLM-12-tk(500) & 0.04 & 7.90 & 1.09 & 0.17 & 0.04 & 93.95 & 98.56 & 4.76 & 48.48 & 8.73\\
    BigLM-12-tp(0.9)  & 0.09 & 11.15 & 2.02 & 0.34 & 0.11 & 92.51 & 97.57 & 4.13 & 34.72 & 7.84\\
    \hline
    BigLM-24-greed & 0.27 & 15.00 & 3.42 & 0.92 & 0.38 & 90.87 & 94.03 & 3.71 & 20.77 & 8.21\\
    BigLM-24-bm(5)  & 0.32 & 14.39 & 4.26 & 1.26 & 0.59 & 85.13 & 89.79 & 4.18 & 19.51 & 6.94\\
    BigLM-24-tk(5) & 0.10 & 12.82 & 2.26 & 0.48 & 0.15 & 90.47 & 95.94 & 3.95 & 27.48 & 7.87\\
    BigLM-24-tk(10) & 0.16 & 11.34 & 1.92 & 0.46 & 0.18 & 91.00 & 96.61 & 3.81 & 29.77 & 8.36\\
    BigLM-24-tk(20) & 0.05 & 10.35 & 1.55 & 0.24 & 0.07 & 92.01 & 97.46 & 3.76 & 33.28 & 8.58\\
    BigLM-24-tk(50) & 0.06 & 9.26 & 1.29 & 0.20 & 0.07 & 92.46 & 97.82 & 3.84 & 38.05 & 9.18\\
    BigLM-24-tk(500) & 0.06 & 7.34 & 0.84 & 0.15 & 0.05 & 93.59 & 98.45 & 4.47 & 47.90 & 9.52\\
    BigLM-24-tp(0.9)  & 0.09 & 10.38 & 1.67 & 0.31 & 0.09 & 91.88 & 97.37 & 3.81 & 33.35 & 8.67\\
    \Xhline{3\arrayrulewidth}
    \end{tabular}}
    \caption{Experimental results on Douban dataset.}
    \label{tab:result-douban}
\end{table*}

\begin{table*}[]
    \centering
    \resizebox{1.9\columnwidth}{!}{
    \begin{tabular}{l|ccccc|cccc|c}
    \Xhline{3\arrayrulewidth}
     \multirow{2}{*}{\textbf{Model}} & \multicolumn{5}{c|}{\textbf{Relevance}} & \multicolumn{4}{c|}{\textbf{Diversity}} & \multirow{2}{*}{\textbf{Length}} \\ \cline{2-10}
    & \textsc{Bleu} & \textsc{B-1} & \textsc{B-2} & \textsc{B-3} & \textsc{B-4} & \textsc{MaDist-1} & \textsc{MaDist-2} & \textsc{MiDist-1} & \textsc{MiDist-2} \\ \hline \hline
    Seq2Seq & 0.71 & 31.28 & 3.61 & 1.56 & 0.99 & 98.23 & 99.69 & 6.77 & 21.88 & 5.53\\
    CVAE& 0.16 & 25.73 & 3.49 & 0.95 & 0.24 & 88.11 & 93.84 & 3.07 & 14.12 & 12.14 \\ 
    MMI  & 0.10 & 23.69 & 2.68 & 0.73 & 0.19 & 96.92 & 98.46 & 2.24 & 8.73 & 8.96\\
    \hline \hline
    Seq2Seq-greedy & 0.66 & 24.51 & 3.24 & 1.36 & 0.83 & 93.56 & 98.83 & 7.53 & 28.82 & 9.10 \\
    Seq2Seq-bm(5) & 0.71 & 31.28 & 3.61 & 1.56 & 0.99 & 98.23 & 99.69 & 6.77 & 21.88 & 5.53\\
    Seq2Seq-tk(5) & 0.37 & 23.06 & 2.64 & 0.87 & 0.56 & 93.04 & 98.98 & 8.85 & 38.43 & 10.54\\
    Seq2Seq-tk(10) & 0.09 & 22.29 & 2.17 & 0.54 & 0.09 & 93.58 & 99.30 & 9.56 & 42.48 & 10.97\\
    Seq2Seq-tk(20) & 0.22 & 21.06 & 1.71 & 0.53 & 0.28 & 93.51 & 99.44 & 10.71 & 47.15 & 11.10\\
    Seq2Seq-tk(50) & 0.14 & 20.71 & 1.82 & 0.41 & 0.14 & 93.74 & 99.56 & 11.83 & 52.02 & 11.12\\
    Seq2Seq-tk(500) & 0.03 & 19.24 & 1.23 & 0.31 & 0.02 & 93.79 & 99.54 & 15.84 & 58.08 & 11.70\\
    Seq2Seq-tp(0.9) & 0.16 & 21.50 & 1.99 & 0.48 & 0.11 & 93.39 & 99.47 & 10.41 & 46.65 & 10.94\\
    \hline
    LM-12-greedy & 0.18 & 18.97 & 2.34 & 0.51 & 0.15 & 73.35 & 84.89 & 1.64 & 8.47 & 21.87\\
    LM-12-bm(5) & 0.21 & 24.13 & 3.21 & 0.83 & 0.26 & 85.16 & 92.41 & 2.35 & 11.05 & 13.61\\
    LM-12-tk(5) & 0.09 & 19.35 & 1.82 & 0.28 & 0.06 & 83.68 & 96.22 & 1.82 & 13.83 & 21.65\\
    LM-12-tk(10) & 0.05 & 18.35 & 1.48 & 0.20 & 0.04 & 85.70 & 97.60 & 1.90 & 16.43 & 22.16\\
    LM-12-tk(20) & 0.03 & 17.37 & 1.27 & 0.16 & 0.02 & 87.08 & 98.22 & 2.04 & 19.49 & 22.46 \\
    LM-12-tk(50) & 0.03 & 16.48 & 1.06 & 0.12 & 0.02 & 88.45 & 98.75 & 2.37 & 23.60 & 22.49 \\
    LM-12-tk(500) & 0.02 & 15.34 & 0.80 & 0.08 & 0.01 & 90.36 & 99.32 & 3.58 & 33.84 & 22.19 \\
    LM-12-tp(0.9) & 0.05 & 17.42 & 1.26 & 0.16 & 0.04 & 87.02 & 98.21 & 2.05 & 19.44 & 22.36\\
    \hline
    LM-12-greedy-$\lambda_1$ & 0.18 & 18.93 & 2.31 & 0.50 & 0.15 & 74.26 & 85.57 & 1.60 & 8.17 & 21.30\\
    LM-12-bm(5)-$\lambda_1$ & 0.21 & 23.26 & 3.07 & 0.80 & 0.26 & 85.27 & 93.22 & 2.10 & 9.62 & 14.38\\
    LM-12-tk(5)-$\lambda_1$ & 0.09 & 19.17 & 1.80 & 0.29 & 0.07 & 83.66 & 96.23 & 1.76 & 13.60 & 21.95\\
    LM-12-tk(10)-$\lambda_1$ & 0.06 & 18.28 & 1.54 & 0.22 & 0.04 & 85.70 & 97.53 & 1.86 & 16.26 & 22.30\\
    LM-12-tk(20)-$\lambda_1$ & 0.05 & 17.33 & 1.30 & 0.17 & 0.04 & 87.13 & 98.29 & 2.02 & 19.29 & 22.53 \\
    LM-12-tk(50)-$\lambda_1$ & 0.03 & 16.25 & 1.04 & 0.13 & 0.02 & 88.40 & 98.76 & 2.33 & 23.37 & 22.71\\
    LM-12-tk(500)-$\lambda_1$ & 0.02 & 15.21 & 0.82 & 0.09 & 0.02 & 90.33 & 99.31 & 3.59 & 33.79 & 22.15\\
    LM-12-tp(0.9)-$\lambda_1$ & 0.04 & 17.31 & 1.26 & 0.16 & 0.03 & 87.12 & 98.24 & 2.01 & 19.18 & 22.58\\
    \hline
    LM-12-greedy-$\lambda_{0.5}$ & 0.18 & 19.16 & 2.35 & 0.50 & 0.15 & 75.05 & 86.15 & 1.58 & 8.06 & 21.12\\
    LM-12-bm(5)-$\lambda_{0.5}$ & 0.22 & 23.51 & 3.13 & 0.81 & 0.27 & 84.93 & 92.81 & 2.16 & 9.94 & 14.24\\
    LM-12-tk(5)-$\lambda_{0.5}$ & 0.09 & 19.19 & 1.79 & 0.29 & 0.06 & 83.76 & 96.24 & 1.73 & 13.34 & 22.04\\
    LM-12-tk(10)-$\lambda_{0.5}$ & 0.05 & 18.14 & 1.47 & 0.20 & 0.03 & 85.61 & 97.54 & 1.83 & 16.12 & 22.60\\
    LM-12-tk(20)-$\lambda_{0.5}$ & 0.03 & 17.15 & 1.25 & 0.15 & 0.02 & 87.15 & 98.31 & 2.00 & 19.23 & 22.63\\
    LM-12-tk(50)-$\lambda_{0.5}$ & 0.03 & 16.39 & 1.06 & 0.12 & 0.02 & 88.41 & 98.74 & 2.32 & 23.29 & 22.83\\
    LM-12-tk(500)-$\lambda_{0.5}$ & 0.02 & 15.19 & 0.80 & 0.09 & 0.02 & 90.36 & 99.31 & 3.52 & 33.45 & 22.35\\
    LM-12-tp(0.9)-$\lambda_{0.5}$ & 0.04 & 17.30 & 1.26 & 0.16 & 0.03 & 87.14 & 98.24 & 1.98 & 19.16 & 22.65\\
    \hline
    LM-24-greedy & 0.15 & 18.04 & 2.14 & 0.44 & 0.11 & 71.42 & 82.96 & 1.14 & 5.93 & 22.66\\
    LM-24-bm(5)  & 0.20 & 22.67 & 2.88 & 0.72 & 0.23 & 83.17 & 90.72 & 1.53 & 7.48 & 14.65\\
    LM-24-tk(5) & 0.08 & 19.38 & 1.80 & 0.27 & 0.05 & 83.66 & 96.10 & 1.47 & 11.79 & 21.43\\
    LM-24-tk(10) & 0.06 & 18.22 & 1.46 & 0.20 & 0.04 & 85.56 & 97.38 & 1.58 & 14.44 & 21.83\\
    LM-24-tk(20) & 0.03 & 17.34 & 1.19 & 0.13 & 0.02 & 87.27 & 98.17 & 1.75 & 17.54 & 21.85 \\
    LM-24-tk(50)& 0.02 & 16.43 & 1.05 & 0.11 & 0.01 & 88.67 & 98.80 & 2.07 & 21.81 & 22.02\\
    LM-24-tk(500) & 0.02 & 15.33 & 0.77 & 0.07 & 0.01 & 90.46 & 99.28 & 3.34 & 32.64 & 21.54\\
    LM-24-tp(0.9) & 0.04 & 17.35 & 1.21 & 0.15 & 0.03 & 87.18 & 98.21 & 1.72 & 17.39 & 21.97\\
    \hline \hline
    BigLM-12-greedy & 0.19 & 19.22 & 2.41 & 0.55 & 0.16 & 74.80 & 85.97 & 2.16 & 10.38 & 21.02  \\
    BigLM-12-bm(5) & 0.21 & 24.18 & 3.32 & 0.86 & 0.25 & 85.81 & 93.01 & 3.55 & 16.11 & 13.37 \\
    BigLM-12-tk(5) & 0.07 & 19.60 & 1.82 & 0.28 & 0.05 & 84.13 & 96.19 & 2.27 & 15.61 & 21.25\\
    BigLM-12-tk(10) & 0.07 & 18.56 & 1.54 & 0.23 & 0.05 & 86.08 & 97.54 & 2.30 & 18.25 & 21.61\\
    BigLM-12-tk(20) & 0.05 & 17.63 & 1.31 & 0.17 & 0.03 & 87.46 & 98.21 & 2.43 & 21.17 & 21.67 \\
    BigLM-12-tk(50) & 0.04 & 16.66 & 1.05 & 0.13 & 0.03 & 88.54 & 98.68 & 2.66 & 25.00 & 21.93\\
    BigLM-12-tk(500) & 0.01 & 15.57 & 0.82 & 0.08 & 0.01 & 90.38 & 99.16 & 3.72 & 34.18 & 21.40 \\
    BigLM-12-tp(0.9) & 0.05 & 17.63 & 1.27 & 0.17 & 0.03 & 87.37 & 98.17 & 2.40 & 21.05 & 21.79 \\
    \hline
    BigLM-24-greedy & 0.20 & 18.68 & 2.36 & 0.53 & 0.15 & 73.15 & 85.16 & 1.97 & 9.65 & 22.24  \\
    BigLM-24-bm(5) & 0.25 & 23.54 & 3.25 & 0.89 & 0.27 & 84.92 & 92.37 & 3.29 & 15.54 & 14.09 \\
    BigLM-24-tk(5)& 0.09 & 19.37 & 1.85 & 0.29 & 0.06 & 83.72 & 96.11 & 2.21 & 15.82 & 21.81\\
    BigLM-24-tk(10) & 0.06 & 18.42 & 1.57 & 0.22 & 0.04 & 85.35 & 97.34 & 2.25 & 18.35 & 22.34\\
    BigLM-24-tk(20)& 0.05 & 17.52 & 1.32 & 0.17 & 0.03 & 86.98 & 98.09 & 2.35 & 21.11 & 22.37 \\
    BigLM-24-tk(50) & 0.04 & 16.63 & 1.13 & 0.15 & 0.03 & 88.33 & 98.61 & 2.64 & 25.17 & 22.38\\
    BigLM-24-tk(500)& 0.02 & 15.50 & 0.85 & 0.09 & 0.02 & 90.06 & 99.13 & 3.68 & 34.19 & 21.96 \\
    BigLM-24-tp(0.9) & 0.04 & 17.56 & 1.35 & 0.18 & 0.03 & 86.98 & 98.10 & 2.35 & 21.14 & 22.29 \\
    \Xhline{3\arrayrulewidth}
    \end{tabular}}
    \caption{Experimental results on Reddit dataset.}
    \label{tab:result-reddit}
\end{table*}

\begin{table*}[]
	\centering
	\resizebox{1.9\columnwidth}{!}{
		\begin{tabular}{l|ccccc|cccc|c}
			\Xhline{3\arrayrulewidth}
			\multirow{2}{*}{\textbf{Model}} & \multicolumn{5}{c|}{\textbf{Relevance}} & \multicolumn{4}{c|}{\textbf{Diversity}} & \multirow{2}{*}{\textbf{Length}} \\ \cline{2-10}
			& \textsc{Bleu} & \textsc{B-1} & \textsc{B-2} & \textsc{B-3} & \textsc{B-4} & \textsc{MaDist-1} & \textsc{MaDist-2} & \textsc{MiDist-1} & \textsc{MiDist-2} \\ \hline \hline
			Seq2Seq & 0.71 & 31.28 & 3.61 & 1.56 & 0.99 & 98.23 & 99.69 & 6.77 & 21.88 & 5.53\\
			CVAE & 5.53 & 38.05 & 13.84 & 9.08 & 5.95 & 95.33 & 99.33 & 7.21 & 29.86 & 6.96 \\ 
			MMI  & 11.91 & 35.53 & 16.31 & 13.55 & 12.07 & 92.78 & 96.03 & 12.30 & 50.78 & 8.36\\
			HRED & 0.67 & 39.27 & 8.62 & 4.37 & 0.99 & 98.47 & 99.57 & 3.78 & 11.89 & 4.67\\
			VHRED & 0.50 & 20.80 & 2.48 & 0.76 & 0.34 & 83.34 & 90.56 & 2.23 & 8.47 & 11.97\\
			VHCR & 0.36 & 23.07 & 2.58 & 1.18 & 0.41 & 86.17 & 92.47 & 2.36 & 8.10 & 10.20 \\
			\hline \hline
			Seq2Seq-greedy & 0.66 & 24.51 & 3.24 & 1.36 & 0.83 & 93.56 & 98.83 & 7.53 & 28.82 & 9.10 \\
			Seq2Seq-bm(5) & 0.71 & 31.28 & 3.61 & 1.56 & 0.99 & 98.23 & 99.69 & 6.77 & 21.88 & 5.53\\
			Seq2Seq-tk(5) & 0.37 & 23.06 & 2.64 & 0.87 & 0.56 & 93.04 & 98.98 & 8.85 & 38.43 & 10.54\\
			Seq2Seq-tk(10) & 0.09 & 22.29 & 2.17 & 0.54 & 0.09 & 93.58 & 99.30 & 9.56 & 42.48 & 10.97\\
			Seq2Seq-tk(20) & 0.22 & 21.06 & 1.71 & 0.53 & 0.28 & 93.51 & 99.44 & 10.71 & 47.15 & 11.10\\
			Seq2Seq-tk(50) & 0.14 & 20.71 & 1.82 & 0.41 & 0.14 & 93.74 & 99.56 & 11.83 & 52.02 & 11.12\\
			Seq2Seq-tk(500) & 0.03 & 19.24 & 1.23 & 0.31 & 0.02 & 93.79 & 99.54 & 15.84 & 58.08 & 11.70\\
			Seq2Seq-tp(0.9) & 0.16 & 21.50 & 1.99 & 0.48 & 0.11 & 93.39 & 99.47 & 10.41 & 46.65 & 10.94\\
			\hline
			LM-12-greedy & 5.67 & 37.96 & 14.94 & 10.19 & 6.12 & 95.09 & 99.28 & 12.86 & 46.06 & 8.15\\
			LM-12-bm(5) & 6.51 & 44.83 & 18.00 & 12.55 & 8.21 & 98.08 & 99.87 & 12.80 & 44.37 & 5.25\\
			LM-12-tk(5) & 3.14 & 30.82 & 9.32 & 5.18 & 3.21 & 94.16 & 99.41 & 13.12 & 50.77 & 9.78\\
			LM-12-tk(10) & 3.29 & 30.05 & 9.01 & 5.71 & 3.51 & 93.99 & 99.30 & 13.48 & 52.81 & 10.25\\
			LM-12-tk(20)& 2.49 & 28.85 & 7.66 & 4.45 & 2.48 & 93.97 & 99.62 & 13.52 & 53.15 & 10.42 \\
			LM-12-tk(50) & 2.27 & 28.22 & 7.23 & 4.29 & 2.36 & 94.35 & 99.60 & 14.92 & 57.01 & 10.50\\
			LM-12-tk(500) & 1.93 & 27.61 & 6.62 & 4.00 & 1.98 & 93.75 & 99.34 & 17.36 & 60.14 & 10.69\\
			LM-12-tp(0.9) & 2.35 & 29.05 & 8.34 & 5.12 & 2.60 & 94.03 & 99.52 & 13.83 & 53.72 & 10.44\\
			\hline
			LM-12-greedy-$\lambda_1$ & 6.21 & 39.56 & 16.48 & 11.55 & 7.11 & 94.76 & 99.23 & 12.60 & 46.27 & 8.31\\
			LM-12-bm(5)-$\lambda_1$ & 6.46 & 45.76 & 18.82 & 13.00 & 7.54 & 97.90 & 99.76 & 12.56 & 42.55 & 5.24\\
			LM-12-tk(5)-$\lambda_1$ & 3.49 & 31.20 & 9.24 & 5.78 & 3.37 & 93.64 & 99.40 & 12.86 & 50.89 & 10.38\\
			LM-12-tk(10)-$\lambda_1$ & 2.58 & 31.17 & 9.10 & 5.39 & 2.75 & 93.87 & 99.54 & 13.23 & 52.84 & 10.19\\
			LM-12-tk(20)-$\lambda_1$ & 2.20 & 28.51 & 8.59 & 4.75 & 2.22 & 93.48 & 99.39 & 13.63 & 55.13 & 11.01 \\
			LM-12-tk(50)-$\lambda_1$ & 2.82 & 28.56 & 9.24 & 5.64 & 2.87 & 93.10 & 99.21 & 14.68 & 57.11 & 11.11\\
			LM-12-tk(500)-$\lambda_1$ & 2.43 & 28.38 & 7.94 & 4.78 & 2.46 & 93.79 & 99.55 & 17.16 & 60.29 & 11.22\\
			LM-12-tp(0.9)-$\lambda_1$ & 2.89 & 30.16 & 8.87 & 5.48 & 2.84 & 93.82 & 99.55 & 13.93 & 53.91 & 10.67\\
			\hline
			LM-12-greedy-$\lambda_{0.5}$ & 6.24 & 37.54 & 15.06 & 10.71 & 6.79 & 94.09 & 99.23 & 11.78 & 43.82 & 8.72\\
			LM-12-bm(5)-$\lambda_{0.5}$ & 5.95 & 44.93 & 17.24 & 11.79 & 7.37 & 97.68 & 99.77 & 12.88 & 42.74 & 5.57\\
			LM-12-tk(5)-$\lambda_{0.5}$ & 2.76 & 30.49 & 8.47 & 4.84 & 2.81 & 93.10 & 99.23 & 12.51 & 49.73 & 10.53\\
			LM-12-tk(10)-$\lambda_{0.5}$ & 2.74 & 29.58 & 8.45 & 4.73 & 2.63 & 93.26 & 99.36 & 12.67 & 51.28 & 11.01\\
			LM-12-tk(20)-$\lambda_{0.5}$ & 2.37 & 27.97 & 7.03 & 4.25 & 2.38 & 93.58 & 99.52 & 13.31 & 53.43 & 11.25\\
			LM-12-tk(50)-$\lambda_{0.5}$& 2.58 & 27.88 & 7.49 & 4.67 & 2.63 & 93.21 & 99.44 & 14.14 & 55.70 & 11.46\\
			LM-12-tk(500)-$\lambda_{0.5}$& 1.91 & 26.12 & 6.37 & 3.68 & 1.96 & 93.34 & 99.51 & 16.58 & 59.99 & 12.09\\
			LM-12-tp(0.9)-$\lambda_{0.5}$ & 2.09 & 29.19 & 8.68 & 4.94 & 2.21 & 93.41 & 99.57 & 13.11 & 53.29 & 11.22\\
			\hline
			LM-24-greedy & 3.63 & 35.90 & 12.14 & 7.62 & 4.43 & 94.90 & 98.94 & 11.17 & 41.27 & 7.86\\
			LM-24-bm(5)  & 3.32 & 42.10 & 13.37 & 8.59 & 4.33 & 98.11 & 99.64 & 11.09 & 37.48 & 4.92\\
			LM-24-tk(5) & 1.94 & 29.59 & 8.00 & 4.36 & 2.09 & 93.42 & 99.17 & 11.68 & 47.66 & 9.82\\
			LM-24-tk(10) & 1.00 & 27.29 & 5.94 & 2.95 & 1.09 & 92.90 & 98.84 & 12.17 & 50.74 & 10.49\\
			LM-24-tk(20) & 1.71 & 27.23 & 6.16 & 3.40 & 1.68 & 93.47 & 99.39 & 12.88 & 53.11 & 10.70 \\
			LM-24-tk(50)& 1.47 & 27.28 & 6.73 & 3.50 & 1.43 & 93.79 & 99.19 & 13.90 & 56.12 & 10.53\\
			LM-24-tk(500) & 0.93 & 24.29 & 5.02 & 2.59 & 1.23 & 93.35 & 99.19 & 16.25 & 60.94 & 11.08\\
			LM-24-tp(0.9) & 1.34 & 26.55 & 6.01 & 3.00 & 1.29 & 93.47 & 99.46 & 12.82 & 53.11 & 10.92\\
			\hline \hline
			BigLM-12-greedy & 6.27 & 34.61 & 11.79 & 8.33 & 6.97 & 94.19 & 97.32 & 13.32 & 48.13 & 8.72  \\
			BigLM-12-bm(5) & 7.56 & 40.93 & 14.56 & 10.95 & 9.12 & 97.55 & 98.11 & 13.70 & 44.31 & 5.87 \\
			BigLM-12-tk(5) & 3.32 & 29.51 & 7.30 & 4.40 & 3.34 & 93.36 & 97.58 & 13.78 & 52.03 & 9.88\\
			BigLM-12-tk(10) & 3.16 & 28.31 & 7.30 & 4.46 & 3.26 & 93.49 & 97.95 & 14.68 & 55.37 & 9.71\\
			BigLM-12-tk(20)& 3.45 & 28.23 & 7.21 & 4.95 & 3.75 & 93.36 & 97.71 & 14.92 & 55.96 & 10.32 \\
			BigLM-12-tk(50) & 3.15 & 28.09 & 6.88 & 4.27 & 3.33 & 93.22 & 97.78 & 15.32 & 57.74 & 10.02\\
			BigLM-12-tk(500)& 2.83 & 27.14 & 6.70 & 4.08 & 2.83 & 93.58 & 98.39 & 17.25 & 59.60 & 10.23 \\
			BigLM-12-tp(0.9) & 3.67 & 28.25 & 7.48 & 5.19 & 3.75 & 94.07 & 99.42 & 14.68 & 56.40 & 10.27 \\
			\hline
			BigLM-24-greedy & 1.71 & 32.16 & 7.25 & 3.81 & 1.89 & 93.85 & 97.99 & 9.05 & 29.87 & 7.58  \\
			BigLM-24-bm(5) & 1.34 & 37.38 & 7.72 & 3.65 & 1.84 & 99.11 & 99.73 & 8.06 & 23.85 & 4.84 \\
			BigLM-24-tk(5) & 0.62 & 27.11 & 4.36 & 2.05 & 0.82 & 93.61 & 99.09 & 11.37 & 43.66 & 9.74\\
			BigLM-24-tk(10) & 0.54 & 25.18 & 3.39 & 1.17 & 0.56 & 93.80 & 99.21 & 12.39 & 47.88 & 10.11\\
			BigLM-24-tk(20)& 0.62 & 24.70 & 3.83 & 1.89 & 0.65 & 93.48 & 99.43 & 12.67 & 51.53 & 10.83 \\
			BigLM-24-tk(50) & 0.30 & 23.74 & 3.07 & 1.32 & 0.46 & 93.74 & 99.39 & 14.24 & 56.29 & 10.66\\
			BigLM-24-tk(500)& 0.19 & 21.93 & 2.93 & 1.04 & 0.19 & 94.28 & 99.57 & 16.28 & 61.74 & 11.33 \\
			BigLM-24-tp(0.9) & 1.11 & 24.86 & 4.45 & 2.24 & 1.23 & 93.00 & 99.13 & 12.83 & 50.91 & 10.78 \\
			\Xhline{3\arrayrulewidth}
	\end{tabular}}
	\caption{Experimental results on DailyDialog dataset.}
	\label{tab:result-daily}
\end{table*}

\begin{table*}[]
    \centering
    \resizebox{1.9\columnwidth}{!}{
    \begin{tabular}{l|ccccc|cccc|c}
    \Xhline{3\arrayrulewidth}
     \multirow{2}{*}{\textbf{Model}} & \multicolumn{5}{c|}{\textbf{Relevance}} & \multicolumn{4}{c|}{\textbf{Diversity}} & \multirow{2}{*}{\textbf{Length}} \\ \cline{2-10}
    & \textsc{Bleu} & \textsc{B-1} & \textsc{B-2} & \textsc{B-3} & \textsc{B-4} & \textsc{MaDist-1} & \textsc{MaDist-2} & \textsc{MiDist-1} & \textsc{MiDist-2} \\ \hline \hline
    Seq2Seq & 0.11 & 14.85 & 1.31 & 0.26 & 0.09 & 97.42 & 99.90 & 6.71 & 24.71 & 7.89\\
    CVAE & 0.04 & 17.20 & 2.04 & 0.28 & 0.04 & 95.22 & 98.71 & 4.65 & 20.80 & 8.48 \\ 
    MMI  & 0.22 & 17.57 & 2.76 & 0.67 & 0.32 & 92.40 & 100.00 & 0.60 & 1.30 & 7.37\\
    HRED & 0.30 & 21.71 & 3.71 & 0.99 & 0.36 & 94.55 & 97.74 & 1.71 & 4.43 & 8.12\\
    VHRED & 0.33 & 18.44 & 2.83 & 0.90 & 0.28 & 87.96 & 95.31 & 1.57 & 4.90 & 9.99\\
    VHCR & 0.30 & 17.32 & 2.43 & 0.65 & 0.21 & 91.16 & 98.15 & 1.89 & 6.18 & 11.10 \\
    \hline \hline
    Seq2Seq-greedy & 0.00 & 14.41 & 1.42 & 0.18 & 0.00 & 94.61 & 99.26 & 6.70 & 28.78 & 10.95 \\
    Seq2Seq-bm(5) & 0.11 & 14.85 & 1.31 & 0.26 & 0.09 & 97.42 & 99.90 & 6.71 & 24.71 & 7.89 \\
    Seq2Seq-tk(5) & 0.07 & 14.50 & 1.38 & 0.21 & 0.03 & 95.21 & 99.70 & 7.74 & 36.27 & 10.58 \\
    Seq2Seq-tk(10) & 0.16 & 13.75 & 1.49 & 0.35 & 0.11 & 95.03 & 99.61 & 8.72 & 40.45 & 10.56 \\
    Seq2Seq-tk(20) & 0.16 & 13.61 & 1.28 & 0.37 & 0.15 & 95.16 & 99.63 & 9.88 & 45.69 & 10.64 \\
    Seq2Seq-tk(50) & 0.02 & 13.32 & 0.91 & 0.08 & 0.01 & 95.57 & 99.68 & 10.91 & 49.35 & 10.58 \\
    Seq2Seq-tk(500) & 0.01 & 13.00 & 0.89 & 0.04 & 0.01 & 95.79 & 99.80 & 15.02 & 57.40 & 10.29 \\
    Seq2Seq-tp(0.9) & 0.07 & 14.04 & 1.30 & 0.17 & 0.06 & 95.34 & 99.63 & 9.49 & 44.72 & 10.67 \\
    \hline
    LM-12-greedy & 0.35 & 19.14 & 3.73 & 0.85 & 0.23 & 91.22 & 98.56 & 9.68 & 37.12 & 10.84\\
    LM-12-bm(5) & 0.21 & 19.94 & 3.44 & 0.71 & 0.16 & 96.67 & 99.61 & 9.56 & 34.09 & 7.72\\
    LM-12-tk(5) & 0.27 & 17.46 & 2.74 & 0.76 & 0.25 & 93.45 & 99.18 & 10.68 & 44.71 & 10.98\\
    LM-12-tk(10) & 0.25 & 16.56 & 2.17 & 0.47 & 0.16 & 94.05 & 99.49 & 11.23 & 47.16 & 10.74\\
    LM-12-tk(20) & 0.16 & 15.97 & 2.24 & 0.51 & 0.16 & 94.16 & 99.50 & 12.15 & 50.40 & 10.89 \\
    LM-12-tk(50) & 0.15 & 15.26 & 1.72 & 0.28 & 0.10 & 94.22 & 99.48 & 12.59 & 52.14 & 10.91 \\
    LM-12-tk(500) & 0.25 & 14.88 & 1.74 & 0.41 & 0.17 & 94.53 & 99.61 & 16.16 & 56.93 & 10.78 \\
    LM-12-tp(0.9) & 0.15 & 16.72 & 2.18 & 0.36 & 0.09 & 94.08 & 99.36 & 11.63 & 49.34 & 10.78\\
    \hline
    LM-12-greedy-$\lambda_1$ & 0.17 & 18.78 & 3.10 & 0.58 & 0.12 & 89.60 & 98.11 & 9.42 & 35.34 & 11.50\\
    LM-12-bm(5)-$\lambda_1$ & 0.76 & 20.57 & 4.10 & 1.39 & 0.66 & 96.07 & 99.64 & 9.12 & 33.29 & 8.15\\
    LM-12-tk(5)-$\lambda_1$ & 0.21 & 17.05 & 2.45 & 0.47 & 0.15 & 92.78 & 99.16 & 10.49 & 43.28 & 11.51\\
    LM-12-tk(10)-$\lambda_1$ & 0.23 & 17.20 & 2.43 & 0.58 & 0.27 & 93.36 & 99.39 & 11.11 & 46.89 & 11.30\\
    LM-12-tk(20)-$\lambda_1$ & 0.05 & 16.62 & 2.04 & 0.43 & 0.06 & 93.38 & 99.43 & 11.68 & 48.51 & 11.61 \\
    LM-12-tk(50)-$\lambda_1$ & 0.15 & 15.78 & 1.83 & 0.45 & 0.13 & 94.05 & 99.49 & 13.15 & 51.80 & 11.27\\
    LM-12-tk(500)-$\lambda_1$ & 0.16 & 15.63 & 1.87 & 0.33 & 0.11 & 94.06 & 99.43 & 15.69 & 55.87 & 11.49\\
    LM-12-tp(0.9)-$\lambda_1$ & 0.08 & 16.40 & 2.15 & 0.28 & 0.06 & 93.29 & 99.48 & 11.69 & 48.04 & 11.52\\
    \hline
    LM-12-greedy-$\lambda_{0.5}$ & 0.22 & 19.19 & 3.12 & 0.70 & 0.14 & 89.82 & 97.87 & 9.26 & 35.46 & 11.50\\
    LM-12-bm(5)-$\lambda_{0.5}$ & 0.30 & 20.80 & 3.45 & 0.92 & 0.22 & 95.83 & 99.34 & 9.58 & 34.31 & 8.22\\
    LM-12-tk(5)-$\lambda_{0.5}$ & 0.14 & 16.92 & 2.58 & 0.48 & 0.13 & 92.46 & 99.19 & 10.98 & 44.42 & 11.49\\
    LM-12-tk(10)-$\lambda_{0.5}$ & 0.09 & 16.33 & 1.93 & 0.41 & 0.16 & 92.95 & 99.02 & 10.84 & 46.77 & 11.46\\
    LM-12-tk(20)-$\lambda_{0.5}$ & 0.15 & 16.68 & 2.32 & 0.52 & 0.12 & 92.98 & 99.36 & 11.79 & 49.08 & 11.51\\
    LM-12-tk(50)-$\lambda_{0.5}$ & 0.07 & 16.32 & 1.83 & 0.30 & 0.04 & 93.95 & 99.38 & 13.09 & 53.65 & 11.41\\
    LM-12-tk(500)-$\lambda_{0.5}$ & 0.03 & 14.75 & 1.44 & 0.24 & 0.02 & 94.02 & 99.61 & 16.42 & 59.15 & 11.53\\
    LM-12-tp(0.9)-$\lambda_{0.5}$ & 0.10 & 16.46 & 2.10 & 0.29 & 0.07 & 93.42 & 99.36 & 12.18 & 50.98 & 11.29\\
    \hline
    LM-24-greedy & 0.35 & 19.14 & 3.73 & 0.85 & 0.23 & 91.22 & 98.56 & 9.68 & 37.12 & 10.84\\
    LM-24-bm(5)  & 0.21 & 19.94 & 3.44 & 0.71 & 0.16 & 96.67 & 99.61 & 9.56 & 34.09 & 7.72\\
    LM-24-tk(5) & 0.20 & 17.77 & 2.71 & 0.66 & 0.18 & 93.29 & 99.22 & 10.80 & 44.20 & 10.90\\
    LM-24-tk(10) & 0.26 & 16.38 & 2.21 & 0.61 & 0.19 & 94.11 & 99.40 & 11.13 & 46.97 & 10.93\\
    LM-24-tk(20) & 0.17 & 15.79 & 1.91 & 0.44 & 0.10 & 94.05 & 99.42 & 11.74 & 49.72 & 11.10 \\
    LM-24-tk(50) & 0.21 & 15.59 & 1.97 & 0.45 & 0.19 & 94.43 & 99.58 & 13.30 & 53.02 & 10.79 \\
    LM-24-tk(500) & 0.11 & 14.71 & 1.67 & 0.37 & 0.07 & 94.27 & 99.46 & 15.79 & 57.61 & 10.82 \\
    LM-24-tp(0.9) & 0.20 & 15.94 & 2.12 & 0.49 & 0.17 & 93.71 & 99.46 & 11.69 & 49.60 & 11.07\\
    \hline \hline
    BigLM-12-greedy & 0.61 & 22.76 & 4.75 & 1.38 & 0.51 & 86.76 & 95.12 & 7.47 & 24.09 & 10.23  \\
    BigLM-12-bm(5) & 0.90 & 24.95 & 5.90 & 2.23 & 1.02 & 96.60 & 99.24 & 7.73 & 22.96 & 7.46 \\
    BigLM-12-tk(5) & 0.43 & 20.02 & 3.55 & 0.93 & 0.36 & 92.11 & 98.59 & 9.10 & 35.54 & 10.84 \\
    BigLM-12-tk(10) & 0.24 & 18.59 & 3.01 & 0.72 & 0.18 & 92.63 & 98.88 & 10.13 & 41.44 & 10.79 \\
    BigLM-12-tk(20) & 0.27 & 18.48 & 2.71 & 0.68 & 0.19 & 93.69 & 99.37 & 10.98 & 46.45 & 10.64 \\
    BigLM-12-tk(50) & 0.18 & 16.81 & 2.03 & 0.36 & 0.11 & 93.43 & 98.87 & 12.07 & 51.59 & 10.74\\
    BigLM-12-tk(500) & 0.16 & 15.86 & 2.10 & 0.46 & 0.10 & 94.38 & 99.08 & 14.90 & 58.31 & 10.67 \\
    BigLM-12-tp(0.9) & 0.20 & 17.87 & 2.57 & 0.56 & 0.21 & 94.01 & 99.28 & 10.83 & 46.17 & 10.65 \\
    \hline
    BigLM-24-greedy & 0.75 & 23.37 & 5.15 & 1.51 & 0.58 & 87.06 & 95.36 & 7.56 & 24.42 & 10.22 \\
    BigLM-24-bm(5) & 1.09 & 25.80 & 6.32 & 2.34 & 1.02 & 96.99 & 99.37 & 7.61 & 23.55 & 7.40 \\
    BigLM-24-tk(5) & 0.39 & 19.36 & 3.40 & 0.95 & 0.38 & 92.42 & 98.61 & 9.05 & 35.51 & 10.69\\
    BigLM-24-tk(10) & 0.22 & 18.84 & 2.74 & 0.66 & 0.17 & 93.40 & 98.98 & 10.18 & 42.14 & 10.66 \\
    BigLM-24-tk(20) & 0.16 & 17.66 & 2.34 & 0.53 & 0.11 & 93.47 & 99.29 & 11.22 & 46.94 & 10.80 \\
    BigLM-24-tk(50) & 0.15 & 17.03 & 1.93 & 0.49 & 0.11 & 93.96 & 98.92 & 12.09 & 51.98 & 10.65 \\
    BigLM-24-tk(500) & 0.05 & 15.57 & 1.97 & 0.34 & 0.03 & 93.61 & 98.99 & 14.85 & 58.27 & 10.96\\
    BigLM-24-tp(0.9) & 0.11 & 17.61 & 2.46 & 0.44 & 0.07 & 93.26 & 99.05 & 10.24 & 46.21 & 10.94 \\
    \Xhline{3\arrayrulewidth}
    \end{tabular}}
    \caption{Experimental results on Persona-Chat dataset.}
    \label{tab:result-persona}
\end{table*}

\section{Examples}
\label{sec:cases}

BigLM-24 models and top-$k$ sampling method are employed to conduct the decoding to generate the following example results. Here $k=32$.  Some of the example queries are extracted from the previous works \cite{zhang2019dialogpt,adiwardana2020towards}.

For each input context (query) $x$, we can repeat it for $b$ times (say $32$) to construct a batch of queries, when we can conduct a parallel decoding approach and $b$ candidate results will be generated within one period. Then we can select the most appropriate one with some ranking strategies such as likelihood or mutual information. The likelihood ranking function is defined as:
\begin{equation}
r = 2^{\frac{1}{n}\sum^{n}_{t=1} \log P(\mathrm{y}_t)}
\label{eql:rank-score}
\end{equation}

\subsection{Results with Ranking Scores}
Table~\ref{fig:reddit-0}$\sim$Table~\ref{fig:reddit-2} and Table~\ref{fig:weibo-0}$\sim$Table~\ref{fig:weibo-2} depict some examples generated by the BigLM-24 models Reddit-Bot and Weibo-Bot respectively. These two bots are both trained for single-turn dialogue generation. The ranking scores are also provided. From the results we can observe that even the candidate results with small ranking scores can still convey some relevance information for the query. It seems that choose the results randomly can also meet the requirement given the query.    

\begin{table*}[!t]
	\centering
	\begin{tabular}{l|p{9cm}|c}
		\hline
		Query & Why do you like the ocean? & Rank Scores \\
		\hline
		Results
& i do n't know how to feel about it , but my mouth is full of seaweed . & 0.3248 \\
& you 'll never know , but i 'd love to see the ocean at some point in the future . & 0.2848 \\
& i like the ocean , i 'll never live like that . & 0.2548 \\
& no , i 'm a girl , i enjoy the ocean ! & 0.2472 \\
& if you ever want a boat that is n't a fish , i 'd be glad to give you one ! & 0.2421 \\
& it 's really not that great . i have plenty of space , but it 's like 10 feet deep . & 0.2399 \\
& that was my favorite part . & 0.2363 \\
& it 's a very nice place to live in anyways & 0.2244 \\
& i actually like the ocean , i thought it was pretty cool and the atmosphere was a pretty good experience . & 0.2233 \\
& lol ! i do not , but it was a good thing i did n't drink and play video games . & 0.2206 \\
& i ca n't really describe what i like on earth besides the ocean . and it 's so clean , it 's nice . & 0.2202 \\
& i really like that ocean , i like the water . & 0.2151 \\
& yes , i agree . that 's why i like a boat . & 0.2111 \\
& there 's no ocean . you need to get a ship . & 0.2054 \\
& i can understand sea . as long as water is in contact with the surface , it 's not a problem . & 0.2022 \\
& lol , well , i 'm not sure what to believe in . they all die at the end . & 0.2012 \\
& you mean the ocean ? you should stop living on the ocean ! & 0.1958 \\
& no , the water is so warm ! & 0.1851 \\
& i like the ocean , just i enjoy drinking the salt from it . & 0.1809 \\
& do you want this to be true or something ? & 0.1790 \\
& what is the sea water ? & 0.1757 \\
& the ocean is the coolest i can get . and the ocean . & 0.1643 \\
& not anime , but i really like it . especially the way the water turns . & 0.1640 \\
& i do too . the ocean is not a ocean . the ocean exists for the people and for all us . & 0.1548 \\
& lol , dude ? lol i like it there ! & 0.1527 \\
& the ocean is amazing for water management purposes , you might want to go out in the open and swim . & 0.1496 \\
& he just likes fish . & 0.1348 \\
& yes , i always liked the sun ! & 0.1346 \\
& yeah xxxx that noise . & 0.1303 \\
& yeah but how else would a sea be a sea ! & 0.1275 \\
& so jealous of america 's coast & 0.1065 \\
& is a lake good ? & 0.1015 \\
		\hline
	\end{tabular}
\caption{\label{fig:reddit-0}
	Single-turn Reddit-Bot example of parallel response generation and ranking. Batch-size is 32. 
}
\vspace{5cm}
\end{table*}

\begin{table*}[!t]
	\centering
	\begin{tabular}{l|p{9cm}|c}
		\hline
		Query & What is the meaning of life ? & Rank Scores \\
		\hline
		Results
& life is the meaning of death . & 0.3715 \\
& the meaning of existence is to live , and the meaning of life is to live . & 0.3565 \\
& it 's a way of life where you do n't lose everything when you do n't lose everything . & 0.3303 \\
& what is life ? what is a word ? & 0.3198 \\
& life is meaningless , it does n't matter what makes you happy & 0.3159 \\
& death is a meaning of life . death is a form of life . life is death . & 0.2996 \\
& death is meaningless . the meaning of death is n't lost on you . & 0.2847 \\
& life is life . it 's fun . & 0.2784 \\
& life 's a joke , yo . & 0.2418 \\
& life is not an act of god . & 0.2389 \\
& the meaning of life : it 's always your thing to do and then there 's your thing to do . & 0.2376 \\
& i guess my question is do you have a good enough reason to think that life is meaningless ? & 0.2336 \\
& i 'm going to need context . i 'm a girl . & 0.2334 \\
& it 's for your life . it 's for your money ! & 0.2279 \\
& life is in the sky , and i 'm born out of this . & 0.1988 \\
& i know that . i feel bad for him . & 0.1972 \\
& the ending is n't very clear to me . it 's very easy to die , it 's pretty easy to go through the effort of trying to live that kind of life after having had your life ruined . & 0.1960 \\
& the same way it 's a good thing you 've made a bad decision since you were born & 0.1894 \\
& i like my life more than my family , so it really is . & 0.1879 \\
& that 's a great answer ! life is what it is , to make you feel alive . & 0.1855 \\
& it makes that meaning meaningful for everybody . i think the fact that the meaning is that no one gives a xxxx about anyone else does n't help . & 0.1839 \\
& life . the universe is what we call life now . & 0.1793 \\
& life is just a thing you do when you are tired . a part of you is not born to love or love the person . & 0.1730 \\
& you can find it on google . & 0.1727 \\
& do n't feel bad . everyone on reddit and everywhere is always saying about you . & 0.1639 \\
& it is life - in the way that it is intended for us , which is why i would like it to remain . & 0.1625 \\
& is it like a fantasy or a fantasy of nothing ? & 0.1560 \\
& i am so happy i can read this sub ! & 0.1407 \\
& nothing . people die . & 0.1359 \\
& death and destruction ! & 0.1304 \\
& i 'd be kind of drunk at dinner and eat my own food . & 0.1268 \\
& just one day when we finally figure out where life is and the future lives us & 0.1000 \\
		\hline
	\end{tabular}
	\caption{\label{fig:reddit-1}
		Another single-turn Reddit-Bot example. 
	}
\end{table*}

\begin{table*}[!t]
	\centering
	\begin{tabular}{l|p{10.5cm}|c}
		\hline
		Query & Coronavirus symptoms: what are they and should I see a doctor? & Rank Scores \\
		\hline
		Results
& what do you mean by symptom ? & 0.3381 \\
& there is no pain . it 's not a sign of adhd , it 's a sign of bad hygiene . & 0.2909 \\
& they are n't vaccinated . & 0.2846 \\
& this is n't one of the symptoms i 'm getting . not sure if that 's a thing in the states , as i do n't know if it 's true in the uk . & 0.2713 \\
& what kind of virus is that ? & 0.2666 \\
& thank you for clarifying , i will give them a try ! & 0.2531 \\
& what 's the reasoning for removing that ? i could n't find it on reddit . & 0.2518 \\
& not to mention , it takes a long time to heal , so the only way to get past it is to eat a whole loaf of bread . & 0.2266 \\
& that would be my best guess . i 'm a little scared since i 'm going to sleep , but the more i think about it the less likely it is to spread to me & 0.2248 \\
& i just checked , and i do n't see the symptoms . i 'll give pills ! & 0.2196 \\
& yeah , that 's what i was saying . i do n't feel like he gave a shit about us being stuck on the  skin for all of a sudden . & 0.2189 \\
& that does n't work either : p & 0.2142 \\
& i do n't see my own face when i look at her . & 0.2130 \\
& the first thing that popped into my head ! i was just about to say : o & 0.2056 \\
& a doctor that does n't need to go to the right subreddit for this to happen . this should be covered . & 0.2006 \\
& i 'll see that doc and see what they think . & 0.2000 \\
& yeah it 's a virus , it 's a bit of an oddity but it 'll cure itself once i get home ! & 0.1990 \\
& just because one person is iskoron , does not mean that this is a normal virus . & 0.1914 \\
& are you referring to a virus or the fact that it is an ingame streaming machine ? or are you referring to the fact this is a server - side exploit and not in real life ? & 0.1870 \\
& no , that 's a really serious infection . & 0.1808 \\
& the thing that confuses me is there is no sign of an adderall sign but it seems like a pretty common sign . & 0.1770 \\
& is this in - game or in the official store ? in - game or in the steam store ? & 0.1768 \\
& i am not sure about how it 's supposed to work , or if it is the cause , but they do say that you need to use oem infused medicine . & 0.1753 \\
& the symptoms are usually not related to the condition . & 0.1724 \\
& how can a person be that stupid for not knowing that they are having any symptoms ? & 0.1605 \\
& i think the medicine 's to slow down the stuttering . & 0.1587 \\
& i was just wondering how it feels . & 0.1503 \\
& what is causing the malfunction ? i see nothing . & 0.1480 \\
& yeah , its the worst , i could n't get my eyes and brain to accept my own disease . if i get one , they look at me like some kind of god - like angel . & 0.1428 \\
& i need this checked my eyes . please tell me how this is caused so i can tell . & 0.1342 \\
& it is an anti virus . the symptoms of their are different . & 0.1325 \\
& i had the same problem , just a friend sent me this error once so it may be wrong . & 0.1204 \\
		\hline
	\end{tabular}
	\caption{\label{fig:reddit-2}
		One more single-turn Reddit-Bot example. 
	}
\end{table*}

\begin{table*}[!t]
	\centering
	\includegraphics[width=2\columnwidth]{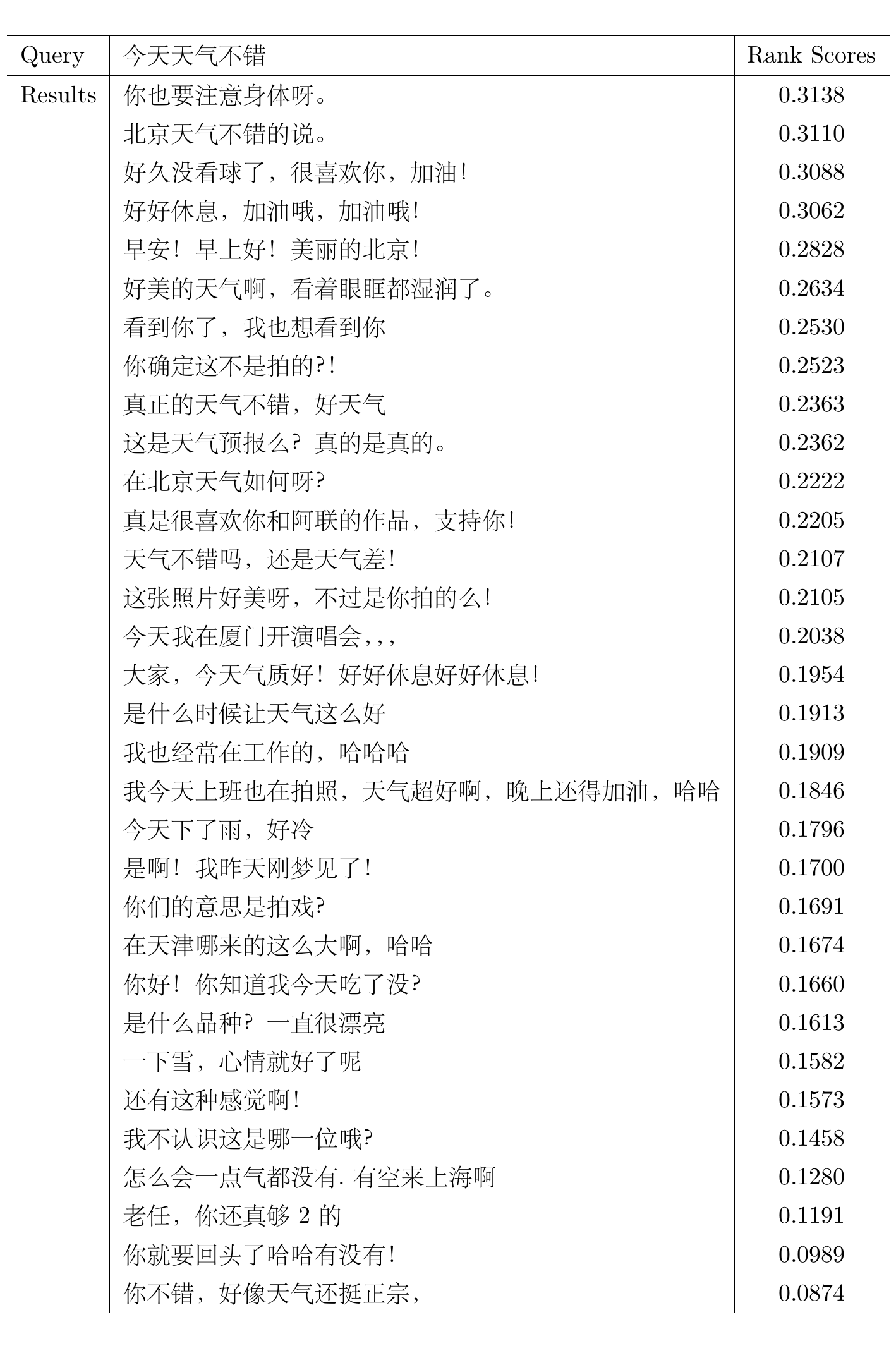}
	\caption{\label{fig:weibo-0}
		Single-turn Weibo-Bot example of parallel response generation and ranking. Batch-size is 32. 
	}
\end{table*}

\begin{table*}[!t]
	\centering
	\includegraphics[width=2\columnwidth]{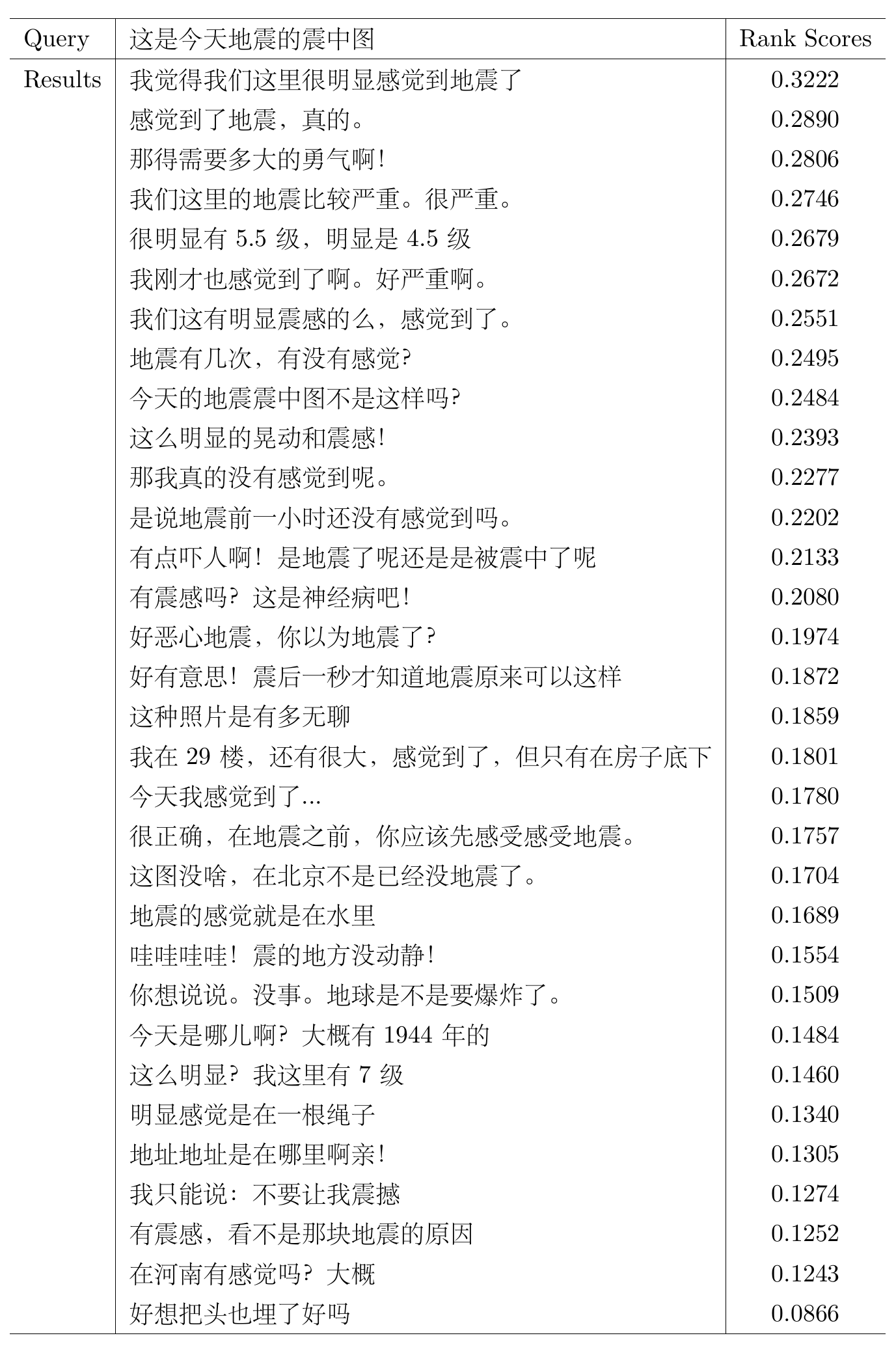}
	\caption{\label{fig:weibo-1}
		Another single-turn Weibo-Bot example. 
	}
\end{table*}

\begin{table*}[!t]
	\centering
	\includegraphics[width=2\columnwidth,height=3\columnwidth]{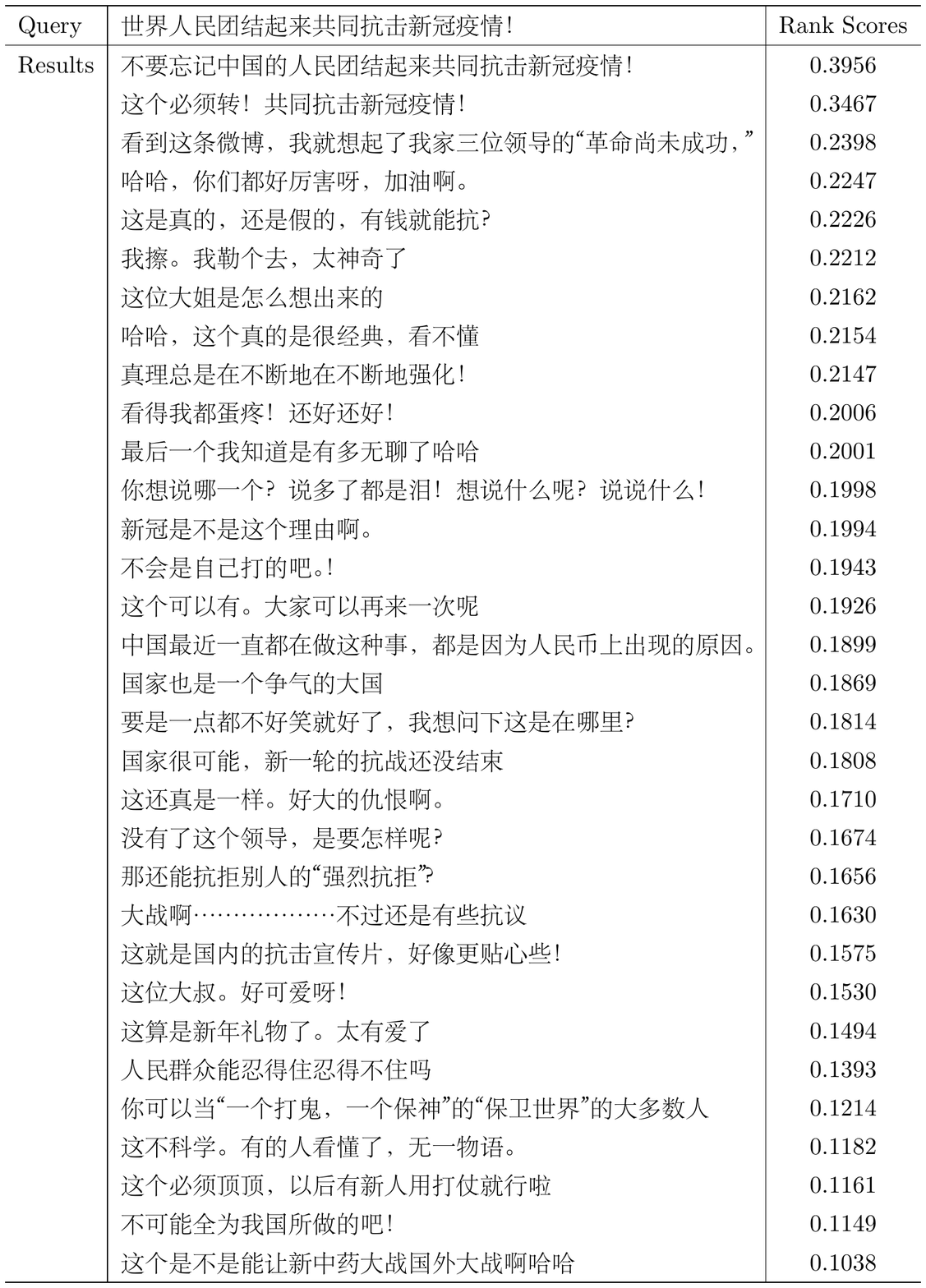}
	\caption{\label{fig:weibo-2}
		One more single-turn Weibo-Bot example. 
	}
\end{table*}

\subsection{Results of Bot-to-Bot  Interactions}
We also conduct bot-to-bot interaction experiments and the example results are described in Table~\ref{fig:daily-0}$\sim$Table~\ref{fig:daily-2}, Table~\ref{fig:reddit-3}, and Table~\ref{fig:douban-0}$\sim$Table~\ref{fig:douban-1} for DailyDialog-Bot, Reddit-Bot, and Douban-Bot respectively. We did not setup the ending single, thus those two-bots can interact infinite turns. We extract the top turns for demonstration. It should be noted that, actually, \textit{no two-bots scenario settings in our work}, we just manage a context memory to generate the next utterance. Therefore, we can observe that the roles for A and B may  interchange with each other during the interactions. And the topic will drift as time goes on. 

\begin{table*}[!t]
	\centering
	\begin{tabular}{l|p{12cm}}
		\hline
		Seed Query & hello \\
		\hline
		Results
& Daily-Bot-A: hello \\
& Daily-Bot-B: hello , this is steven . i'm calling to confirm the date of my appointment . \\
& Daily-Bot-A: it's july 5th . \\
& Daily-Bot-B: congratulations ! i hope you can make it as soon as possible . \\
& Daily-Bot-A: i'm afraid so . i have an appointment at three pm on july 10th . \\
& Daily-Bot-B: ok , i will see you then . \\
& Daily-Bot-A: very well , thank you for your co-operation . \\
& Daily-Bot-B: i am looking forward to hearing from you . \\
& Daily-Bot-A: good-bye . \\
& Daily-Bot-B: see you then . \\
		\hline
	\end{tabular}
\caption{\label{fig:daily-0}
	An example of interactions between two multi-turn DailyDialog-Bots.  
}
\end{table*}

\begin{table*}[!t]
	\centering
	\begin{tabular}{l|p{12cm}}
		\hline
		Seed Query & hello \\
		\hline
		Results
& Daily-Bot-A: hello \\
& Daily-Bot-B: hello , this is lucy from abc company . i ’ ll be glad to meet you . \\
& Daily-Bot-A: good to meet you , lucy . \\
& Daily-Bot-B: i ’ m calling to inform you that we have arranged an interview for this accountant position at 2 pm this thursday afternoon . please come on time . \\
& Daily-Bot-A: ok , thank you . \\
& Daily-Bot-B: you are welcome . \\
& Daily-Bot-A: i hope i can apply for the position as soon as possible . \\
& Daily-Bot-B: yes , please . please send your resume to our department . \\
& Daily-Bot-A: ok , thank you . \\
& Daily-Bot-B: you ’ re welcome . \\
& Daily-Bot-A: it ’ s my pleasure having your firm ’ s christmas tree delivered to you . \\
& Daily-Bot-B: it ’ s my pleasure . \\
& Daily-Bot-A: we will send someone to your office in a few minutes . see you then . \\
& Daily-Bot-B: that ’ s all right . \\
& Daily-Bot-A: that ’ s all right . \\
& Daily-Bot-B: don ’ t worry . i ’ ll be straight home as soon as i have all of our christmas ready . \\
& Daily-Bot-A: thanks a much . \\
& Daily-Bot-B: you ’ re welcome . \\
& Daily-Bot-A: i ’ m glad you could come so soon . \\
& Daily-Bot-B: i appreciate your help . \\
& Daily-Bot-A: don ’ t worry . it ’ s nothing . \\
& Daily-Bot-B: i ’ m glad that we have a big christmas dinner . \\
& Daily-Bot-A: i hope so . \\
& Daily-Bot-B: i hope so , too . \\
& Daily-Bot-A: you ’ re welcome . \\
& Daily-Bot-B: i hope so , too . \\
& Daily-Bot-A: bye-bye . \\
		\hline
	\end{tabular}
	\caption{\label{fig:daily-1}
		Another example of interactions between two multi-turn DailyDialog-Bots.  
	}
\end{table*}

\begin{table*}[!t]
	\centering
	\begin{tabular}{l|p{12cm}}
		\hline
		Seed Query & Why do you like the ocean? \\
		\hline
		Results
& Daily-Bot-A: why do you like the ocean? \\
& Daily-Bot-B: it's pleasant while the sun is shining . \\
& Daily-Bot-A: i like the feeling then that the world changes so quickly . \\
& Daily-Bot-B: i agree . people have personal styles and qualities that suit their needs . \\
& Daily-Bot-A: i agree.it seems that many people like the seafood . \\
& Daily-Bot-B: well , what i would like is some sole . \\
& Daily-Bot-A: ok . i will bring it back to our shop tonight . \\
& Daily-Bot-B: sure , sir . have a nice day . \\
& Daily-Bot-A: thank you.thanks for inviting me . \\
& Daily-Bot-B: you are welcome . \\
& Daily-Bot-A: don't mention it . \\
& Daily-Bot-B: it's been a long time since i had sole . \\
& Daily-Bot-A: next time , please . \\
& Daily-Bot-B: that's all right . \\
& Daily-Bot-A: than you would have liked , sir . \\
& Daily-Bot-B: it's be nice . \\
& Daily-Bot-A: then , shall we shorten the distance ? \\
& Daily-Bot-B: let me see ... just a minute . \\
& Daily-Bot-A: you're not quite right . \\
& Daily-Bot-B: i suppose i could , sir . \\
& Daily-Bot-A: yes , of course . that will be great . \\
& Daily-Bot-B: and for our bottles , please . \\
& Daily-Bot-A: yes , sir . we'll do it business through the atlas . \\
& Daily-Bot-B: thank you very much . \\
& Daily-Bot-A: we're sure your wine would be appreciated . \\
& Daily-Bot-B: i'm glad to hear that . \\
& Daily-Bot-A: would you please fill out this form ? here you are . \\
& Daily-Bot-B: now let me check . \\
& Daily-Bot-A: by the way , please open this bottle for me . everything will be paid for in advance . \\
& Daily-Bot-B: certainly , sir . \\
& Daily-Bot-A: here is the forty dollars . \\
& Daily-Bot-B: wait a moment , please . \\
& Daily-Bot-A: ( after a while ) i am sorry you mustn't enjoy the service . \\
& Daily-Bot-B: that's all right . \\
& Daily-Bot-A: here is the money . \\
& Daily-Bot-B: thank you very much . i hope you take it back . \\
& Daily-Bot-A: it's my pleasure . \\
& Daily-Bot-B: come back again a little later . \\
& Daily-Bot-A: i've never paid much attention to the service in my life . \\
& Daily-Bot-B: it's normal . i appreciate the time you've prepared for it . \\
& Daily-Bot-A: i appreciate the time you've prepared . \\
& Daily-Bot-B: it's my pleasure . \\
& Daily-Bot-A: call me back any time you need help . \\
& Daily-Bot-B: i appreciate the time . \\
& Daily-Bot-A: don't mention it . \\
& Daily-Bot-B: it's my pleasure . \\
& Daily-Bot-A: our friendship is mutual . \\
& Daily-Bot-B: you're welcome . \\
& Daily-Bot-A: it's a pleasure . \\
& Daily-Bot-B: i'll glad to do well . \\
		\hline
	\end{tabular}
	\caption{\label{fig:daily-2}
		One more example of interactions between two multi-turn DailyDialog-Bots.  
	}
\end{table*}

\begin{table*}[!t]
	\centering
	\begin{tabular}{l|p{13cm}}
		\hline
		Seed Query & What is the meaning of life? \\
		\hline
		Results
& Reddit-Bot-A: what is the meaning of life? \\
& Reddit-Bot-B: the meaning of life ? the purpose of human existence ? like the purpose of love or the purpose of happiness ? \\
& Reddit-Bot-A: life is not something you can live . \\
& Reddit-Bot-B: that 's why everything on the world ends when you die , and that is because you live forever . you cannot die when you die . life happens and you live forever . \\
& Reddit-Bot-A: i just do n't understand how you could think that . \\
& Reddit-Bot-B: this is why i do n't even consider myself a happy person . \\
& Reddit-Bot-A: do you think life could ever be any sort of happy ending ? or just be a happy happy ending ? \\
& Reddit-Bot-B: because you die with a single soul ? and you die on a tangled web you ca n't move on from ? \\
& Reddit-Bot-A: i 'm not sure . i think death will be better when everyone is able to support a human being without a child , especially in certain circumstances . \\
& Reddit-Bot-B: he 'll be fine , he 'll die as long as he keeps his soul around . \\
& Reddit-Bot-A: it 's the same as people complaining how they die or people complaining how they die because they got too worked up about things . \\
& Reddit-Bot-B: i ca n't believe someone said that . someone has to be a complete xxx to you to not try . \\
& Reddit-Bot-A: i 'm assuming he was just trying to say there 's a difference between death and cruelty \\
& Reddit-Bot-B: i mean its pretty obvious that he made a mistake and you 're just mad because of it \\
& Reddit-Bot-A: nah , he just makes up thing to sound cool as hell . \\
& Reddit-Bot-B: no i understand it 's a joke but it 's sad and offensive that someone would think it 's an okay thing to do to somebody that makes fun of them . \\
& Reddit-Bot-A: if you like his joke then you probably do n't like the way it works in real life . i would think this might be the kind of person that 'd get his audience to laugh when they see you . \\
& Reddit-Bot-B: oh , i see what you mean . the thing being fun is probably a joke . i 'm sorry this happened to you . \\
& Reddit-Bot-A: what other thing happens to you ? i ca n't tell what else you 're talking about . \\
& Reddit-Bot-B: we 're a bunch of kids all about fun . sometimes funny thing happens to people too . it 's funny . it makes me uncomfortable \\
& Reddit-Bot-A: just cause i have to share my story does n't mean i do n't have too much fun in life . i like sharing that story without having to share every part of it with someone else . \\
& Reddit-Bot-B: my brother has this issue too ! he tells everyone he knows every week . \\
& Reddit-Bot-A: i 've seen some guys go through the same thing . i know that 's what it sounds like so i 'm just trying to share my story with the good ones . \\
& Reddit-Bot-B: yea that 's why i like it . i can get the story out of the way sometimes but my best friend told him that i was going to tell him that if we talked to him . \\
& Reddit-Bot-A: yeah , it 's kind of hard to get what you want when everyone loves you so much so much . i 'm sorry for you . \\
& Reddit-Bot-B: i 'm very sorry . it 's not hard at all . i have a long lasting relationship and we 've been together for a long time \\
& Reddit-Bot-A: no worries . if you ever need advice , feel free to message me and i 'll answer any questions i have . \\
		\hline
	\end{tabular}
	\caption{\label{fig:reddit-3}
		An example of interactions between two single-turn Reddit-Bots.  
	}
\end{table*}

\begin{table*}[!t]
	\centering
	\includegraphics[width=1.7\columnwidth]{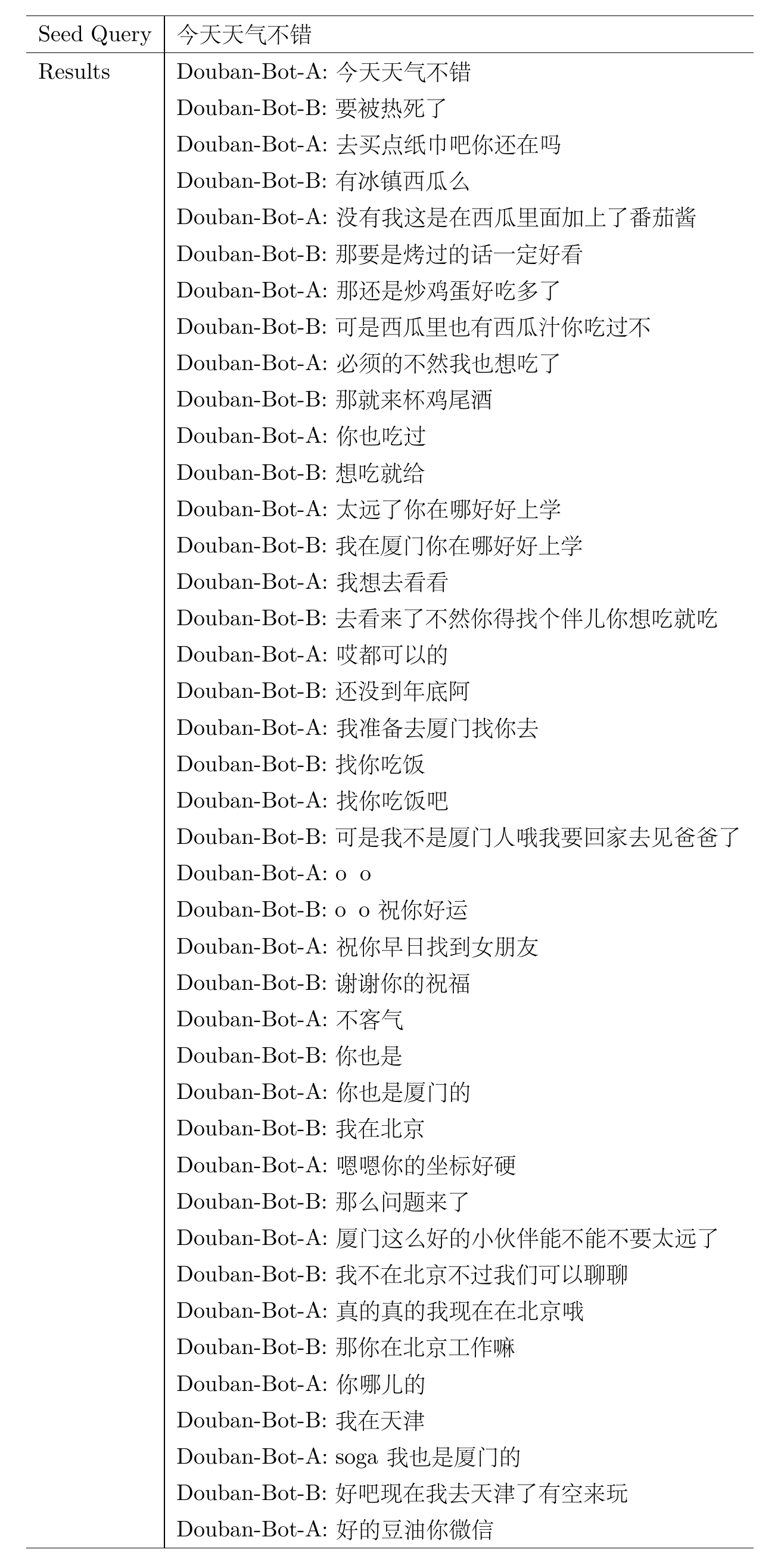}
	\caption{\label{fig:douban-0}
		Example of interactions between two multi-turn Douban-Bots.  
	}
\end{table*}

\begin{table*}[!t]
	\centering
	\includegraphics[width=1.65\columnwidth]{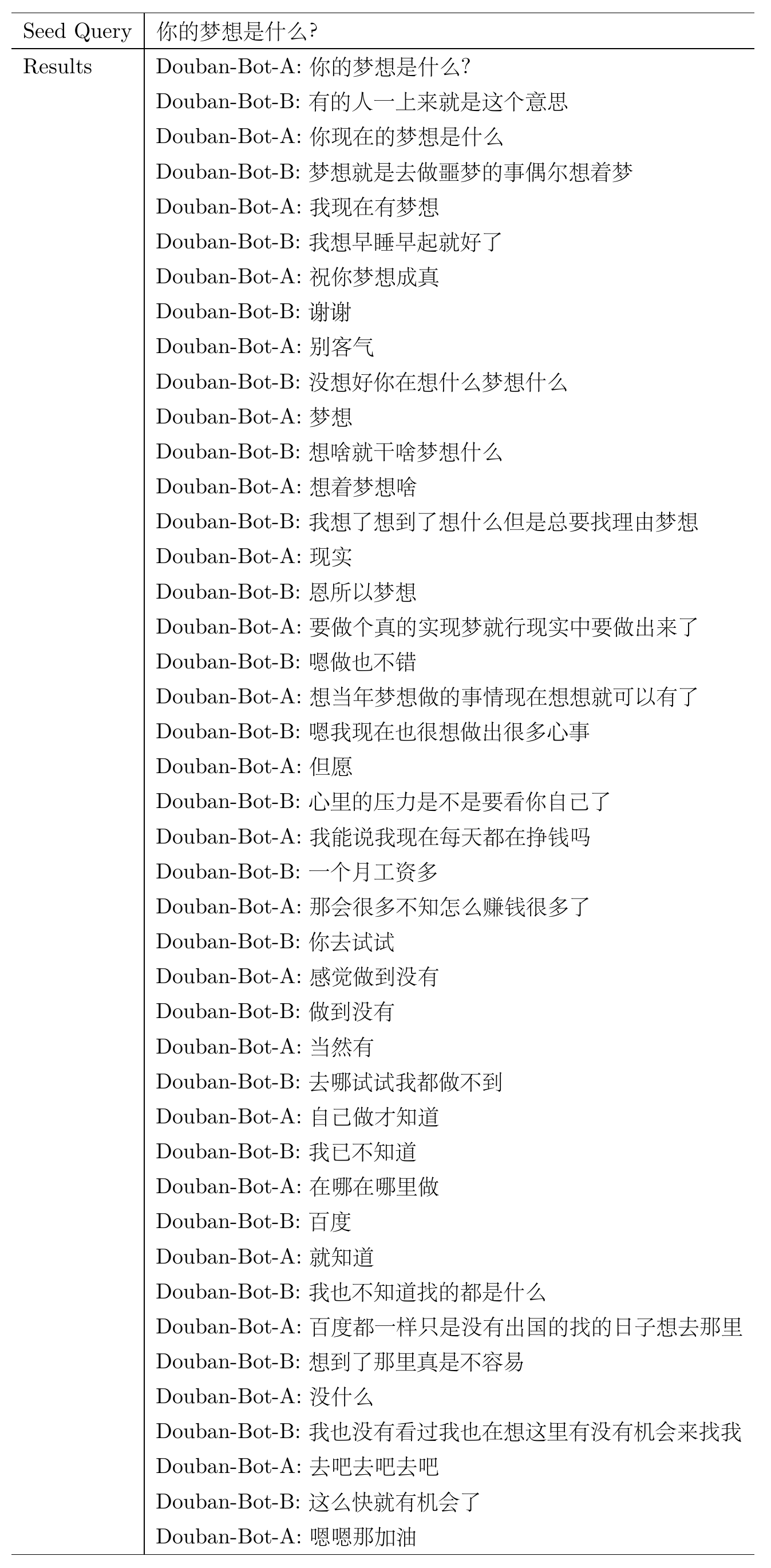}
	\caption{\label{fig:douban-1}
		Another example of interactions between two multi-turn Douban-Bots.  
	}
\end{table*}

\section{Conclusions}
\label{sec:cons}

We present an empirical investigation of pre-trained Transformer-based auto-regressive language models for the task of open-domain dialogue generation.  Training paradigm of pre-training and fine-tuning is employed to conduct the parameter learning. Various of decoding strategies are employed to conduct the response text generation.  
Extensive experiments are conducted on the typical single-turn and multi-turn dialogue datasets. Detailed numbers of automatic evaluation metrics on relevance and diversity of the generated results for the languages models as well as the baseline approaches are reported.

Nevertheless, some serious issues also exist in the generated results, such as the grammatical issue and the topic drift problem. More seriously, sometimes the bots will generate contrafactual or offensive content.  Therefore, better model structures, training paradigms, and decoding strategies need to be investigated and constructed in the future.

\bibliography{acl2020}

\begin{thebibliography}{62}
\expandafter\ifx\csname natexlab\endcsname\relax\def\natexlab#1{#1}\fi

\bibitem[{Adiwardana et~al.(2020)Adiwardana, Luong, So, Hall, Fiedel,
  Thoppilan, Yang, Kulshreshtha, Nemade, Lu et~al.}]{adiwardana2020towards}
Daniel Adiwardana, Minh-Thang Luong, David~R So, Jamie Hall, Noah Fiedel, Romal
  Thoppilan, Zi~Yang, Apoorv Kulshreshtha, Gaurav Nemade, Yifeng Lu, et~al.
  2020.
\newblock Towards a human-like open-domain chatbot.
\newblock \emph{arXiv preprint arXiv:2001.09977}.

\bibitem[{Bahdanau et~al.(2015)Bahdanau, Cho, and Bengio}]{bahdanau2015neural}
Dzmitry Bahdanau, Kyunghyun Cho, and Yoshua Bengio. 2015.
\newblock Neural machine translation by jointly learning to align and
  translate.
\newblock In \emph{3rd International Conference on Learning Representations,
  ICLR 2015}.

\bibitem[{Bi et~al.(2019)Bi, Gao, Liu, and Shi}]{bi2019fine}
Wei Bi, Jun Gao, Xiaojiang Liu, and Shuming Shi. 2019.
\newblock Fine-grained sentence functions for short-text conversation.
\newblock In \emph{Proceedings of the 57th Annual Meeting of the Association
  for Computational Linguistics}, pages 3984--3993.

\bibitem[{Chan et~al.(2019)Chan, Li, Yang, Chen, Hu, Zhao, and
  Yan}]{chan2019modeling}
Zhangming Chan, Juntao Li, Xiaopeng Yang, Xiuying Chen, Wenpeng Hu, Dongyan
  Zhao, and Rui Yan. 2019.
\newblock Modeling personalization in continuous space for response generation
  via augmented wasserstein autoencoders.
\newblock In \emph{Proceedings of the 2019 Conference on Empirical Methods in
  Natural Language Processing and the 9th International Joint Conference on
  Natural Language Processing (EMNLP-IJCNLP)}, pages 1931--1940.

\bibitem[{Cho et~al.(2014)Cho, van Merri{\"e}nboer, Gulcehre, Bahdanau,
  Bougares, Schwenk, and Bengio}]{cho2014learning}
Kyunghyun Cho, Bart van Merri{\"e}nboer, Caglar Gulcehre, Dzmitry Bahdanau,
  Fethi Bougares, Holger Schwenk, and Yoshua Bengio. 2014.
\newblock Learning phrase representations using rnn encoder--decoder for
  statistical machine translation.
\newblock In \emph{Proceedings of the 2014 Conference on Empirical Methods in
  Natural Language Processing (EMNLP)}, pages 1724--1734.

\bibitem[{Cuay{\'a}huitl(2017)}]{cuayahuitl2017simpleds}
Heriberto Cuay{\'a}huitl. 2017.
\newblock Simpleds: A simple deep reinforcement learning dialogue system.
\newblock In \emph{Dialogues with social robots}, pages 109--118. Springer.

\bibitem[{Devlin et~al.(2019)Devlin, Chang, Lee, and
  Toutanova}]{devlin2019bert}
Jacob Devlin, Ming-Wei Chang, Kenton Lee, and Kristina Toutanova. 2019.
\newblock Bert: Pre-training of deep bidirectional transformers for language
  understanding.
\newblock In \emph{Proceedings of the 2019 Conference of the North American
  Chapter of the Association for Computational Linguistics: Human Language
  Technologies, Volume 1 (Long and Short Papers)}, pages 4171--4186.

\bibitem[{Dhingra et~al.(2017)Dhingra, Li, Li, Gao, Chen, Ahmad, and
  Deng}]{dhingra2017towards}
Bhuwan Dhingra, Lihong Li, Xiujun Li, Jianfeng Gao, Yun-Nung Chen, Faisal
  Ahmad, and Li~Deng. 2017.
\newblock Towards end-to-end reinforcement learning of dialogue agents for
  information access.
\newblock In \emph{Proceedings of the 55th Annual Meeting of the Association
  for Computational Linguistics (Volume 1: Long Papers)}, pages 484--495.

\bibitem[{Dinan et~al.(2018)Dinan, Roller, Shuster, Fan, Auli, and
  Weston}]{dinan2018wizard}
Emily Dinan, Stephen Roller, Kurt Shuster, Angela Fan, Michael Auli, and Jason
  Weston. 2018.
\newblock Wizard of wikipedia: Knowledge-powered conversational agents.
\newblock \emph{arXiv preprint arXiv:1811.01241}.

\bibitem[{Dong et~al.(2019)Dong, Yang, Wang, Wei, Liu, Wang, Gao, Zhou, and
  Hon}]{dong2019unified}
Li~Dong, Nan Yang, Wenhui Wang, Furu Wei, Xiaodong Liu, Yu~Wang, Jianfeng Gao,
  Ming Zhou, and Hsiao-Wuen Hon. 2019.
\newblock Unified language model pre-training for natural language
  understanding and generation.
\newblock In \emph{Advances in Neural Information Processing Systems}, pages
  13042--13054.

\bibitem[{Fan et~al.(2018)Fan, Lewis, and Dauphin}]{fan2018hierarchical}
Angela Fan, Mike Lewis, and Yann Dauphin. 2018.
\newblock Hierarchical neural story generation.
\newblock In \emph{Proceedings of the 56th Annual Meeting of the Association
  for Computational Linguistics (Volume 1: Long Papers)}, pages 889--898.

\bibitem[{Gao et~al.(2019{\natexlab{a}})Gao, Galley, Li et~al.}]{gao2019neural}
Jianfeng Gao, Michel Galley, Lihong Li, et~al. 2019{\natexlab{a}}.
\newblock Neural approaches to conversational ai.
\newblock \emph{Foundations and Trends{\textregistered} in Information
  Retrieval}, 13(2-3):127--298.

\bibitem[{Gao et~al.(2019{\natexlab{b}})Gao, Bi, Liu, Li, and
  Shi}]{gao2019generating}
Jun Gao, Wei Bi, Xiaojiang Liu, Junhui Li, and Shuming Shi. 2019{\natexlab{b}}.
\newblock Generating multiple diverse responses for short-text conversation.
\newblock In \emph{Proceedings of the AAAI Conference on Artificial
  Intelligence}, volume~33, pages 6383--6390.

\bibitem[{Gao et~al.(2019{\natexlab{c}})Gao, Bi, Liu, Li, Zhou, and
  Shi}]{gao2019discrete}
Jun Gao, Wei Bi, Xiaojiang Liu, Junhui Li, Guodong Zhou, and Shuming Shi.
  2019{\natexlab{c}}.
\newblock A discrete cvae for response generation on short-text conversation.
\newblock In \emph{Proceedings of the 2019 Conference on Empirical Methods in
  Natural Language Processing and the 9th International Joint Conference on
  Natural Language Processing (EMNLP-IJCNLP)}, pages 1898--1908.

\bibitem[{Ghazvininejad et~al.(2018)Ghazvininejad, Brockett, Chang, Dolan, Gao,
  Yih, and Galley}]{ghazvininejad2018knowledge}
Marjan Ghazvininejad, Chris Brockett, Ming-Wei Chang, Bill Dolan, Jianfeng Gao,
  Wen-tau Yih, and Michel Galley. 2018.
\newblock A knowledge-grounded neural conversation model.
\newblock In \emph{Thirty-Second AAAI Conference on Artificial Intelligence}.

\bibitem[{Gu et~al.(2018)Gu, Cho, Ha, and Kim}]{gu2018dialogwae}
Xiaodong Gu, Kyunghyun Cho, Jung-Woo Ha, and Sunghun Kim. 2018.
\newblock Dialogwae: Multimodal response generation with conditional
  wasserstein auto-encoder.
\newblock \emph{arXiv preprint arXiv:1805.12352}.

\bibitem[{Hochreiter and Schmidhuber(1997)}]{hochreiter1997long}
Sepp Hochreiter and J{\"u}rgen Schmidhuber. 1997.
\newblock Long short-term memory.
\newblock \emph{Neural computation}, 9(8):1735--1780.

\bibitem[{Holtzman et~al.(2019)Holtzman, Buys, Forbes, and
  Choi}]{holtzman2019curious}
Ari Holtzman, Jan Buys, Maxwell Forbes, and Yejin Choi. 2019.
\newblock The curious case of neural text degeneration.
\newblock \emph{arXiv preprint arXiv:1904.09751}.

\bibitem[{Huang et~al.(2018)Huang, Zaiane, Trabelsi, and
  Dziri}]{huang2018automatic}
Chenyang Huang, Osmar~R Zaiane, Amine Trabelsi, and Nouha Dziri. 2018.
\newblock Automatic dialogue generation with expressed emotions.
\newblock In \emph{Proceedings of the 2018 Conference of the North American
  Chapter of the Association for Computational Linguistics: Human Language
  Technologies, Volume 2 (Short Papers)}, pages 49--54.

\bibitem[{Ke et~al.(2018)Ke, Guan, Huang, and Zhu}]{ke2018generating}
Pei Ke, Jian Guan, Minlie Huang, and Xiaoyan Zhu. 2018.
\newblock Generating informative responses with controlled sentence function.
\newblock In \emph{Proceedings of the 56th Annual Meeting of the Association
  for Computational Linguistics (Volume 1: Long Papers)}, pages 1499--1508.

\bibitem[{Kingma and Ba(2014)}]{kingma2014adam}
Diederik~P Kingma and Jimmy Ba. 2014.
\newblock Adam: A method for stochastic optimization.
\newblock \emph{arXiv preprint arXiv:1412.6980}.

\bibitem[{Kingma and Welling(2013)}]{kingma2013auto}
Diederik~P Kingma and Max Welling. 2013.
\newblock Auto-encoding variational bayes.
\newblock \emph{arXiv preprint arXiv:1312.6114}.

\bibitem[{Koehn(2004)}]{koehn2004pharaoh}
Philipp Koehn. 2004.
\newblock Pharaoh: a beam search decoder for phrase-based statistical machine
  translation models.
\newblock In \emph{Conference of the Association for Machine Translation in the
  Americas}, pages 115--124. Springer.

\bibitem[{Li et~al.(2016{\natexlab{a}})Li, Galley, Brockett, Gao, and
  Dolan}]{li2016diversity}
Jiwei Li, Michel Galley, Chris Brockett, Jianfeng Gao, and Bill Dolan.
  2016{\natexlab{a}}.
\newblock A diversity-promoting objective function for neural conversation
  models.
\newblock In \emph{Proceedings of the 2016 Conference of the North American
  Chapter of the Association for Computational Linguistics: Human Language
  Technologies}, pages 110--119.

\bibitem[{Li et~al.(2016{\natexlab{b}})Li, Galley, Brockett, Spithourakis, Gao,
  and Dolan}]{li2016persona}
Jiwei Li, Michel Galley, Chris Brockett, Georgios Spithourakis, Jianfeng Gao,
  and Bill Dolan. 2016{\natexlab{b}}.
\newblock A persona-based neural conversation model.
\newblock In \emph{Proceedings of the 54th Annual Meeting of the Association
  for Computational Linguistics (Volume 1: Long Papers)}, pages 994--1003.

\bibitem[{Li et~al.(2016{\natexlab{c}})Li, Monroe, Ritter, Jurafsky, Galley,
  and Gao}]{li2016deep}
Jiwei Li, Will Monroe, Alan Ritter, Dan Jurafsky, Michel Galley, and Jianfeng
  Gao. 2016{\natexlab{c}}.
\newblock Deep reinforcement learning for dialogue generation.
\newblock In \emph{Proceedings of the 2016 Conference on Empirical Methods in
  Natural Language Processing}, pages 1192--1202.

\bibitem[{Li et~al.(2017{\natexlab{a}})Li, Monroe, Shi, Jean, Ritter, and
  Jurafsky}]{li2017adversarial}
Jiwei Li, Will Monroe, Tianlin Shi, S{\'e}bastien Jean, Alan Ritter, and Dan
  Jurafsky. 2017{\natexlab{a}}.
\newblock Adversarial learning for neural dialogue generation.
\newblock In \emph{Proceedings of the 2017 Conference on Empirical Methods in
  Natural Language Processing}, pages 2157--2169.

\bibitem[{Li et~al.(2019{\natexlab{a}})Li, Chen, Ren, Chen, Tu, and
  Ma}]{li2019empgan}
Qintong Li, Hongshen Chen, Zhaochun Ren, Zhumin Chen, Zhaopeng Tu, and Jun Ma.
  2019{\natexlab{a}}.
\newblock Empgan: Multi-resolution interactive empathetic dialogue generation.
\newblock \emph{arXiv preprint arXiv:1911.08698}.

\bibitem[{Li et~al.(2019{\natexlab{b}})Li, Li, Bi, Liu, and
  Lam}]{li2019relevance}
Xin Li, Piji Li, Wei Bi, Xiaojiang Liu, and Wai Lam. 2019{\natexlab{b}}.
\newblock Relevance-promoting language model for short-text conversation.
\newblock \emph{arXiv preprint arXiv:1911.11489}.

\bibitem[{Li et~al.(2017{\natexlab{b}})Li, Su, Shen, Li, Cao, and
  Niu}]{li2017dailydialog}
Yanran Li, Hui Su, Xiaoyu Shen, Wenjie Li, Ziqiang Cao, and Shuzi Niu.
  2017{\natexlab{b}}.
\newblock Dailydialog: A manually labelled multi-turn dialogue dataset.
\newblock In \emph{Proceedings of the Eighth International Joint Conference on
  Natural Language Processing (Volume 1: Long Papers)}, pages 986--995.

\bibitem[{Liu and Singh(2004)}]{liu2004conceptnet}
Hugo Liu and Push Singh. 2004.
\newblock Conceptnet—a practical commonsense reasoning tool-kit.
\newblock \emph{BT technology journal}, 22(4):211--226.

\bibitem[{Madotto et~al.(2019)Madotto, Lin, Wu, and
  Fung}]{madotto2019personalizing}
Andrea Madotto, Zhaojiang Lin, Chien-Sheng Wu, and Pascale Fung. 2019.
\newblock Personalizing dialogue agents via meta-learning.
\newblock In \emph{Proceedings of the 57th Annual Meeting of the Association
  for Computational Linguistics}, pages 5454--5459.

\bibitem[{Mazare et~al.(2018)Mazare, Humeau, Raison, and
  Bordes}]{mazare2018training}
Pierre-Emmanuel Mazare, Samuel Humeau, Martin Raison, and Antoine Bordes. 2018.
\newblock Training millions of personalized dialogue agents.
\newblock In \emph{Proceedings of the 2018 Conference on Empirical Methods in
  Natural Language Processing}, pages 2775--2779.

\bibitem[{Olabiyi and Mueller(2019)}]{olabiyi2019multi}
Oluwatobi Olabiyi and Erik~T Mueller. 2019.
\newblock Multi-turn dialogue response generation with autoregressive
  transformer models.
\newblock \emph{arXiv preprint arXiv:1908.01841}.

\bibitem[{Pan et~al.(2019)Pan, Bai, Wang, Zhou, and Liu}]{pan2019improving}
Zhufeng Pan, Kun Bai, Yan Wang, Lianqiang Zhou, and Xiaojiang Liu. 2019.
\newblock Improving open-domain dialogue systems via multi-turn incomplete
  utterance restoration.
\newblock In \emph{Proceedings of the 2019 Conference on Empirical Methods in
  Natural Language Processing and the 9th International Joint Conference on
  Natural Language Processing (EMNLP-IJCNLP)}, pages 1824--1833.

\bibitem[{Park et~al.(2018)Park, Cho, and Kim}]{park2018hierarchical}
Yookoon Park, Jaemin Cho, and Gunhee Kim. 2018.
\newblock A hierarchical latent structure for variational conversation
  modeling.
\newblock In \emph{Proceedings of the 2018 Conference of the North American
  Chapter of the Association for Computational Linguistics: Human Language
  Technologies, Volume 1 (Long Papers)}, pages 1792--1801.

\bibitem[{Qiu et~al.(2019)Qiu, Li, Bi, Zhao, and Yan}]{qiu2019training}
Lisong Qiu, Juntao Li, Wei Bi, Dongyan Zhao, and Rui Yan. 2019.
\newblock Are training samples correlated? learning to generate dialogue
  responses with multiple references.
\newblock In \emph{Proceedings of the 57th Annual Meeting of the Association
  for Computational Linguistics}, pages 3826--3835.

\bibitem[{Radford et~al.(2019{\natexlab{a}})Radford, Wu, Amodei, Amodei, Clark,
  Brundage, and Sutskever}]{radford2019better}
Alec Radford, Jeffrey Wu, Dario Amodei, Daniela Amodei, Jack Clark, Miles
  Brundage, and Ilya Sutskever. 2019{\natexlab{a}}.
\newblock Better language models and their implications.
\newblock \emph{OpenAI Blog https://openai. com/blog/better-language-models}.

\bibitem[{Radford et~al.(2019{\natexlab{b}})Radford, Wu, Child, Luan, Amodei,
  and Sutskever}]{radford2019language}
Alec Radford, Jeffrey Wu, Rewon Child, David Luan, Dario Amodei, and Ilya
  Sutskever. 2019{\natexlab{b}}.
\newblock Language models are unsupervised multitask learners.
\newblock \emph{OpenAI Blog}, 1(8):9.

\bibitem[{Rashkin et~al.(2019)Rashkin, Smith, Li, and
  Boureau}]{rashkin2019towards}
Hannah Rashkin, Eric~Michael Smith, Margaret Li, and Y-Lan Boureau. 2019.
\newblock Towards empathetic open-domain conversation models: A new benchmark
  and dataset.
\newblock In \emph{Proceedings of the 57th Annual Meeting of the Association
  for Computational Linguistics}, pages 5370--5381.

\bibitem[{See et~al.(2019)See, Pappu, Saxena, Yerukola, and
  Manning}]{see2019massively}
Abigail See, Aneesh Pappu, Rohun Saxena, Akhila Yerukola, and Christopher~D
  Manning. 2019.
\newblock Do massively pretrained language models make better storytellers?
\newblock In \emph{Proceedings of the 23rd Conference on Computational Natural
  Language Learning (CoNLL)}, pages 843--861.

\bibitem[{Sennrich et~al.(2016)Sennrich, Haddow, and
  Birch}]{sennrich2016neural}
Rico Sennrich, Barry Haddow, and Alexandra Birch. 2016.
\newblock Neural machine translation of rare words with subword units.
\newblock In \emph{Proceedings of the 54th Annual Meeting of the Association
  for Computational Linguistics (Volume 1: Long Papers)}, pages 1715--1725.

\bibitem[{Serban et~al.(2016)Serban, Sordoni, Bengio, Courville, and
  Pineau}]{serban2016building}
Iulian~V Serban, Alessandro Sordoni, Yoshua Bengio, Aaron Courville, and Joelle
  Pineau. 2016.
\newblock Building end-to-end dialogue systems using generative hierarchical
  neural network models.
\newblock In \emph{Thirtieth AAAI Conference on Artificial Intelligence}.

\bibitem[{Serban et~al.(2017)Serban, Sordoni, Lowe, Charlin, Pineau, Courville,
  and Bengio}]{serban2017hierarchical}
Iulian~Vlad Serban, Alessandro Sordoni, Ryan Lowe, Laurent Charlin, Joelle
  Pineau, Aaron Courville, and Yoshua Bengio. 2017.
\newblock A hierarchical latent variable encoder-decoder model for generating
  dialogues.
\newblock In \emph{Thirty-First AAAI Conference on Artificial Intelligence}.

\bibitem[{Shang et~al.(2015)Shang, Lu, and Li}]{shang2015neural}
Lifeng Shang, Zhengdong Lu, and Hang Li. 2015.
\newblock Neural responding machine for short-text conversation.
\newblock In \emph{Proceedings of the 53rd Annual Meeting of the Association
  for Computational Linguistics and the 7th International Joint Conference on
  Natural Language Processing (Volume 1: Long Papers)}, pages 1577--1586.

\bibitem[{So et~al.(2019)So, Le, and Liang}]{so2019evolved}
David So, Quoc Le, and Chen Liang. 2019.
\newblock The evolved transformer.
\newblock In \emph{International Conference on Machine Learning}, pages
  5877--5886.

\bibitem[{Sordoni et~al.(2015)Sordoni, Galley, Auli, Brockett, Ji, Mitchell,
  Nie, Gao, and Dolan}]{sordoni2015neural}
Alessandro Sordoni, Michel Galley, Michael Auli, Chris Brockett, Yangfeng Ji,
  Margaret Mitchell, Jian-Yun Nie, Jianfeng Gao, and Bill Dolan. 2015.
\newblock A neural network approach to context-sensitive generation of
  conversational responses.
\newblock In \emph{Proceedings of the 2015 Conference of the North American
  Chapter of the Association for Computational Linguistics: Human Language
  Technologies}, pages 196--205.

\bibitem[{Vaswani et~al.(2017)Vaswani, Shazeer, Parmar, Uszkoreit, Jones,
  Gomez, Kaiser, and Polosukhin}]{vaswani2017attention}
Ashish Vaswani, Noam Shazeer, Niki Parmar, Jakob Uszkoreit, Llion Jones,
  Aidan~N Gomez, {\L}ukasz Kaiser, and Illia Polosukhin. 2017.
\newblock Attention is all you need.
\newblock In \emph{Advances in neural information processing systems}, pages
  5998--6008.

\bibitem[{Vinyals and Le(2015)}]{vinyals2015neural}
Oriol Vinyals and Quoc Le. 2015.
\newblock A neural conversational model.
\newblock \emph{arXiv preprint arXiv:1506.05869}.

\bibitem[{Wu et~al.(2017)Wu, Wu, Xing, Zhou, and Li}]{wu2017sequential}
Yu~Wu, Wei Wu, Chen Xing, Ming Zhou, and Zhoujun Li. 2017.
\newblock Sequential matching network: A new architecture for multi-turn
  response selection in retrieval-based chatbots.
\newblock In \emph{Proceedings of the 55th Annual Meeting of the Association
  for Computational Linguistics (Volume 1: Long Papers)}, pages 496--505.

\bibitem[{Xu et~al.(2020)Xu, Li, Yang, Ren, Ren, Chen, and Ma}]{xu2020neural}
Minghong Xu, Piji Li, Haoran Yang, Pengjie Ren, Zhaochun Ren, Zhumin Chen, and
  Jun Ma. 2020.
\newblock A neural topical expansion framework for unstructured
  persona-oriented dialogue generation.
\newblock \emph{arXiv preprint arXiv:2002.02153}.

\bibitem[{Xu et~al.(2017)Xu, Liu, Wang, Sun, Wang, Wang, and Qi}]{xu2017neural}
Zhen Xu, Bingquan Liu, Baoxun Wang, Cheng-Jie Sun, Xiaolong Wang, Zhuoran Wang,
  and Chao Qi. 2017.
\newblock Neural response generation via gan with an approximate embedding
  layer.
\newblock In \emph{Proceedings of the 2017 Conference on Empirical Methods in
  Natural Language Processing}, pages 617--626.

\bibitem[{Young et~al.(2018)Young, Cambria, Chaturvedi, Zhou, Biswas, and
  Huang}]{young2018augmenting}
Tom Young, Erik Cambria, Iti Chaturvedi, Hao Zhou, Subham Biswas, and Minlie
  Huang. 2018.
\newblock Augmenting end-to-end dialogue systems with commonsense knowledge.
\newblock In \emph{Thirty-Second AAAI Conference on Artificial Intelligence}.

\bibitem[{Zhang et~al.(2018{\natexlab{a}})Zhang, Dinan, Urbanek, Szlam, Kiela,
  and Weston}]{zhang2018personalizing}
Saizheng Zhang, Emily Dinan, Jack Urbanek, Arthur Szlam, Douwe Kiela, and Jason
  Weston. 2018{\natexlab{a}}.
\newblock Personalizing dialogue agents: I have a dog, do you have pets too?
\newblock In \emph{Proceedings of the 56th Annual Meeting of the Association
  for Computational Linguistics (Volume 1: Long Papers)}, pages 2204--2213.

\bibitem[{Zhang et~al.(2018{\natexlab{b}})Zhang, Galley, Gao, Gan, Li,
  Brockett, and Dolan}]{zhang2018generating}
Yizhe Zhang, Michel Galley, Jianfeng Gao, Zhe Gan, Xiujun Li, Chris Brockett,
  and Bill Dolan. 2018{\natexlab{b}}.
\newblock Generating informative and diverse conversational responses via
  adversarial information maximization.
\newblock In \emph{Advances in Neural Information Processing Systems}, pages
  1810--1820.

\bibitem[{Zhang et~al.(2019)Zhang, Sun, Galley, Chen, Brockett, Gao, Gao, Liu,
  and Dolan}]{zhang2019dialogpt}
Yizhe Zhang, Siqi Sun, Michel Galley, Yen-Chun Chen, Chris Brockett, Xiang Gao,
  Jianfeng Gao, Jingjing Liu, and Bill Dolan. 2019.
\newblock Dialogpt: Large-scale generative pre-training for conversational
  response generation.
\newblock \emph{arXiv preprint arXiv:1911.00536}.

\bibitem[{Zhao et~al.(2017)Zhao, Zhao, and Eskenazi}]{zhao2017learning}
Tiancheng Zhao, Ran Zhao, and Maxine Eskenazi. 2017.
\newblock Learning discourse-level diversity for neural dialog models using
  conditional variational autoencoders.
\newblock In \emph{Proceedings of the 55th Annual Meeting of the Association
  for Computational Linguistics (Volume 1: Long Papers)}, pages 654--664.

\bibitem[{Zheng et~al.(2019)Zheng, Zhang, Mao, and Huang}]{zheng2019pre}
Yinhe Zheng, Rongsheng Zhang, Xiaoxi Mao, and Minlie Huang. 2019.
\newblock A pre-training based personalized dialogue generation model with
  persona-sparse data.
\newblock \emph{arXiv preprint arXiv:1911.04700}.

\bibitem[{Zhou et~al.(2018{\natexlab{a}})Zhou, Huang, Zhang, Zhu, and
  Liu}]{zhou2018emotional}
Hao Zhou, Minlie Huang, Tianyang Zhang, Xiaoyan Zhu, and Bing Liu.
  2018{\natexlab{a}}.
\newblock Emotional chatting machine: Emotional conversation generation with
  internal and external memory.
\newblock In \emph{Thirty-Second AAAI Conference on Artificial Intelligence}.

\bibitem[{Zhou et~al.(2018{\natexlab{b}})Zhou, Young, Huang, Zhao, Xu, and
  Zhu}]{zhou2018commonsense}
Hao Zhou, Tom Young, Minlie Huang, Haizhou Zhao, Jingfang Xu, and Xiaoyan Zhu.
  2018{\natexlab{b}}.
\newblock Commonsense knowledge aware conversation generation with graph
  attention.
\newblock In \emph{Proceedings of the 27th International Joint Conference on
  Artificial Intelligence}, pages 4623--4629.

\bibitem[{Zhou et~al.(2018{\natexlab{c}})Zhou, Gao, Li, and
  Shum}]{zhou2018design}
Li~Zhou, Jianfeng Gao, Di~Li, and Heung-Yeung Shum. 2018{\natexlab{c}}.
\newblock The design and implementation of xiaoice, an empathetic social
  chatbot.
\newblock \emph{Computational Linguistics}, (Just Accepted):1--62.

\bibitem[{Zhu et~al.(2015)Zhu, Kiros, Zemel, Salakhutdinov, Urtasun, Torralba,
  and Fidler}]{zhu2015aligning}
Yukun Zhu, Ryan Kiros, Rich Zemel, Ruslan Salakhutdinov, Raquel Urtasun,
  Antonio Torralba, and Sanja Fidler. 2015.
\newblock Aligning books and movies: Towards story-like visual explanations by
  watching movies and reading books.
\newblock In \emph{Proceedings of the IEEE international conference on computer
  vision}, pages 19--27.

\end{thebibliography}
\bibliographystyle{acl_natbib}

\end{document}